\documentclass{article}

\PassOptionsToPackage{numbers}{natbib}
 \usepackage[preprint]{neurips_2026}

\usepackage[utf8]{inputenc} 
\usepackage[T1]{fontenc}    
\usepackage{hyperref}       
\usepackage{url}            
\usepackage{booktabs}       
\usepackage{amsfonts}       
\usepackage{nicefrac}       
\usepackage{microtype}      
\usepackage{xcolor}         
\usepackage{booktabs}
\usepackage{multirow}
\usepackage{amsmath}
\usepackage{enumitem}
\usepackage[pdftex]{graphicx}
\usepackage{amssymb}

\usepackage{algorithm}
\usepackage{algpseudocode}
\usepackage{threeparttable}
\usepackage{multirow}
\usepackage{amsmath}
\usepackage{amssymb}
\DeclareMathOperator{\Clamp}{Clamp}
\DeclareMathOperator{\sg}{sg}

\usepackage{tcolorbox}
\tcbuselibrary{breakable}
\tcbuselibrary{breakable, skins} 

\definecolor{mytitlebg}{RGB}{68, 114, 198}
\definecolor{mybodybg}{RGB}{207, 213, 234}
\newtcolorbox{promptbox}[1][]{
  colback=mybodybg,                  
  colframe=mytitlebg,                
  colbacktitle=mytitlebg,
  arc=2mm,                  
  boxrule=0.5pt,            
  breakable,                
  fontupper=\small\sffamily,
  left=2mm, right=2mm,      
  top=2mm, bottom=2mm,      
  #1                        
}

\title{Efficient LLM Reasoning via Variational Posterior Guidance with Efficiency Awareness}

\author{%
  Zizhao Chen \quad Yuying Li \quad Siting Lin \quad Lianxi Wang\thanks{Corresponding author.} \\
  Guangdong University of Foreign Studies, Guangzhou, China.\\
  \texttt\{chenzizhao,wanglianxi\}@gdufs.edu.cn \\
}

\begin{document}

\maketitle

\begin{abstract}

Although large language models rely on chain-of-thought for complex reasoning, the overthinking phenomenon severely degrades inference efficiency. Existing reinforcement learning methods compress reasoning chains by designing elaborate reward functions, which renders high-quality samples extremely sparse in the exploration space and creates a sampling bottleneck for the prior policy. Inspired by cognitive science, we theoretically prove that a posterior distribution guided by reference answers achieves higher expected utility than the prior distribution, thus capable of breaking through the sampling bottleneck of high-quality samples. However, the posterior distribution is unavailable during inference. To this end, we formalize efficient reasoning as a variational inference problem and introduce an efficiency-aware evidence lower bound as the theoretical foundation. Based on this, we propose the VPG-EA framework. It adopts a parameter-shared dual-stream architecture to instantiate both the posterior distribution and the prior policy; after filtering out pseudo-efficient paths via cross-view evaluation, it unidirectionally transfers the posterior's efficient patterns to the prior policy through variational distillation. Experiments on DeepSeek-R1-Distill-Qwen-1.5B and 7B scales demonstrate that VPG-EA improves the comprehensive efficiency metric $\varepsilon^3$ by 8.73\% and 12.37\% over the strongest baselines on each model size, respectively.

\end{abstract}

\section{Introduction}

As large language models (LLMs) enter the System 2 reasoning era \citep{DBLP:journals/pami/ZhangLZZLYXZCZYDGSL26}, reinforcement learning (RL) algorithms significantly improve their mathematical and logical reasoning capabilities through process or outcome supervision signals \citep{lightman2024lets,yu2025dapo}. However, existing RL algorithms \citep{luo2025opruner,yeo2025demystifying,shen2025dast} often encourage detailed intermediate steps to achieve higher accuracy. This causes models to learn an implicit strategy. They tend to rely on lengthy reasoning paths to maximize rewards \citep{DBLP:journals/tmlr/SuiCWZZYLWZZCH25}. This strategy causes severe overthinking when generalizing to simple problems \citep{chen2024not}. Even for simple queries, models tend to generate many unnecessary redundant steps. This trades computational cost for accuracy.

To alleviate overthinking, existing studies mainly design length penalty rewards to compress reasoning chains \citep{luo2025opruner,yeo2025demystifying,shen2025dast}. However, these methods face a severe sample efficiency bottleneck: trajectories that are simultaneously correct and concise occupy an extremely sparse efficient manifold. Traditional RL algorithms, such as GRPO \citep{shao2024deepseekmath} and PPO \citep{schulman2017proximal}, rely on random exploration from the prior policy, which rarely reaches this narrow manifold. If sampling fails to cover these efficient samples, gradients cannot update effectively regardless of the reward function design, creating a critical hurdle for traditional RL in balancing accuracy and efficiency.

Recent studies introduce Variational Inference (VI) \citep{kingma2013auto}, utilizing reference-answer-guided posterior distributions to overcome sampling barriers on extremely difficult tasks \citep{zhou2026variational,lin2025ravr}. Given that VI offers a principled pathway to break through sampling bottlenecks, we formalize the efficient reasoning problem through the lens of variational inference. Using this theory, we derive new mathematical boundaries and establish an efficiency-aware evidence lower bound (ELBO) as a guiding heuristic.

\begin{figure}[htbp]
  \centering
  \includegraphics[width=0.8\linewidth]{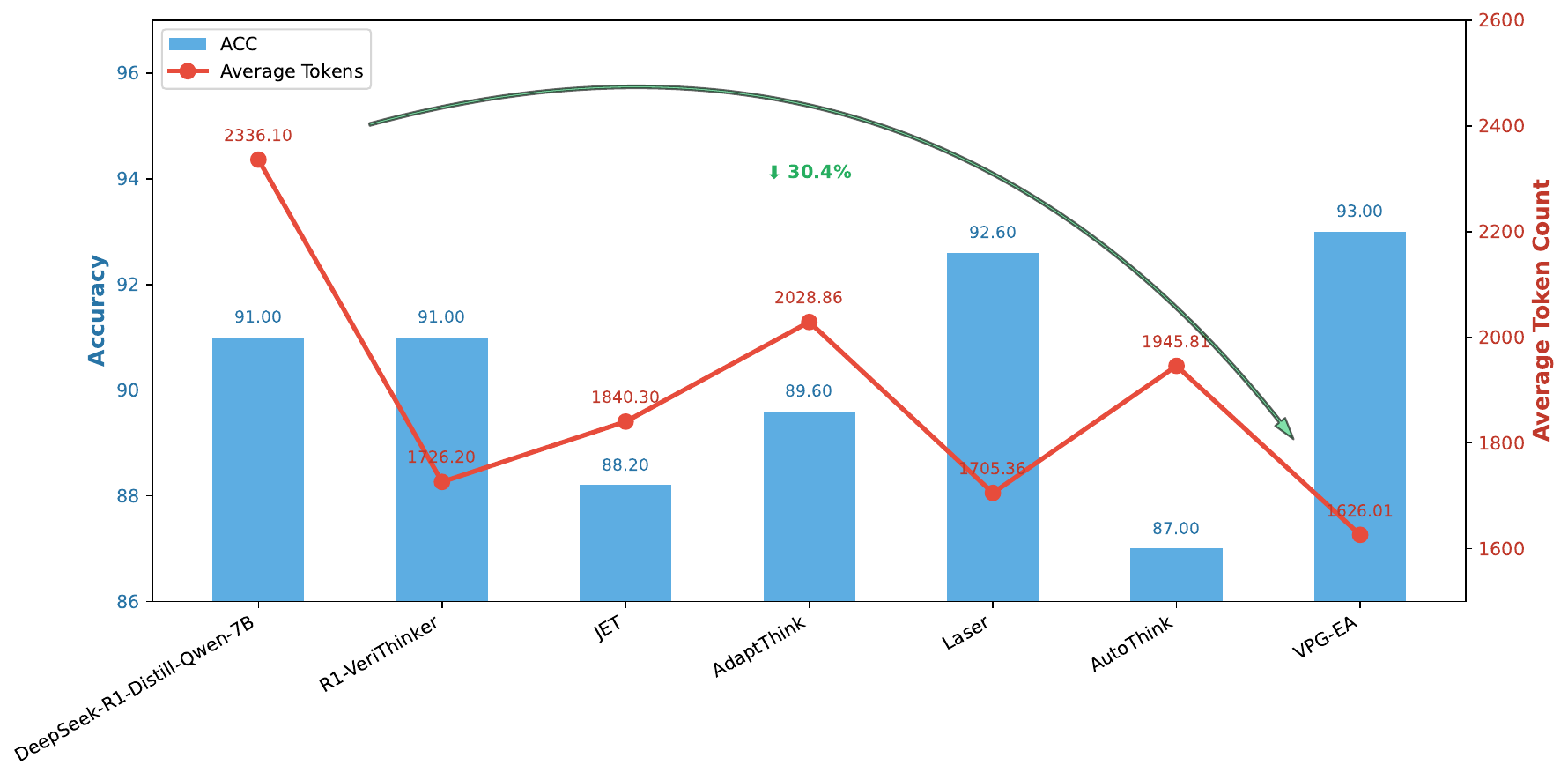}
  \caption{Performance comparison of VPG-EA with various baselines on MATH-500.}
  \label{fig:scatter_plot}
\end{figure}

Built upon this structural foundation, we propose the VPG-EA (Variational Posterior Guidance and Efficiency Awareness) framework, applying VI theory to LLM reasoning efficiency optimization for the first time. Specifically, we strictly prove the expected utility advantage of the reference-guided posterior. To instantiate our ELBO-inspired objective, VPG-EA employs a dual-stream sampling architecture equipped with cross-view evaluation to filter out pseudo-efficient paths and alleviate reward hacking. Through dynamic advantage gating and variational distillation, it smoothly transfers the verified efficient patterns from the posterior to the prior policy. As shown in Figure \ref{fig:scatter_plot}, this mechanism thoroughly resolves the sampling challenge of traditional RL, achieving a dual improvement in both the reasoning accuracy and efficiency of LLMs. The main contributions of this paper include:
\begin{itemize}[leftmargin=*]
    \item We theoretically prove that a reference-answer-guided posterior distribution has an expected utility advantage over the prior distribution.
    
    \item We propose VPG-EA. It locates the efficient manifold through posterior guidance. It utilizes variational distillation to achieve robust knowledge transfer to the prior policy. 
    
    \item Experiments on 1.5B and 7B models show that VPG-EA improves the comprehensive efficiency metric $\varepsilon^3$ by 8.73\% and 12.37\%, respectively. It reduces token consumption by over 30\%.
\end{itemize}

\section{Related Work}

\paragraph{Efficiency Optimization via Reward Engineering}
With the emergence of overthinking in long CoT reasoning, recent studies primarily regulate reasoning length through RL reward engineering. One category designs refined length-penalty functions to guide shortening within standard RL frameworks \citep{luo2025opruner,yeo2025demystifying,arora2025training}. Another explores difficulty-adaptive or multi-stage pruning, such as difficulty-budget calibration \citep{shen2025dast}, decoupling reasoning patterns from generation \citep{fang2025thinkless}, and dynamic length scheduling \citep{hou2026thinkprune,tu2025learning,aggarwal2025l} and self-aware boundary adaptation \citep{chen2026aware}. While these methods effectively shorten average CoT length, severe length penalization exacerbates the extreme sparsity of samples that are simultaneously correct and concise. This causes traditional RL algorithms relying on random sampling to face severe sample efficiency bottlenecks during exploration, making it extremely difficult to reach and learn the narrow efficient reasoning manifold.

\paragraph{Reference-answer-guided Exploration}
To alleviate exploration difficulties in high-difficulty tasks, an emerging line of research introduces reference answers as guidance signals. Early methods utilize correct answers to construct positive samples via iterative self-training (STaR) \citep{zelikman2024star} or dynamic clue injection (Nudging) \citep{chen2026nudging} during RL. Recent studies formalize this as probabilistic inference: ExPO \citep{zhou2025expo} generates high-likelihood paths within trust regions, while RAVR \citep{lin2025ravr} and concurrent work \citep{zhou2026variational} model reasoning under VI frameworks, treating thought trajectories as latent variables for optimization. Although these reference-answer-guided methods successfully break sampling barriers for high-difficulty tasks, they are fundamentally constrained to capability enhancement. Their core inductive bias assumes any trajectory deriving the correct answer is optimal, entirely ignoring the endogenous tendency of LLMs to trade steps for accuracy \citep{DBLP:journals/tmlr/SuiCWZZYLWZZCH25}. In stark contrast, our work explicitly targets reasoning efficiency by formalizing it as a utility maximization problem under length-aware constraints. Rather than merely porting existing variational objectives, we adapt the standard ELBO framework as a heuristic by incorporating an efficiency-aware term. 

\section{Preliminaries and Problem Definition}

\subsection{Problem Formalization}
\label{sec:problem_formalization}

Given a verifiable reasoning task dataset $D=\{(x_i,y^*_i)\}_{i=1}^N$ containing $N$ question-answer pairs, where $x$ represents the input question and $y^*$ represents the correct answer. LLMs reason from question $x$ to answer $y$ via a chain of thought (latent variable $z$). The joint probability distribution can be factorized as:
\begin{equation}
P_\theta(y,z|x)=P_\theta(y|x,z)\cdot \pi_\theta(z|x),
\label{eq:eq1}
\end{equation}
here, $\pi_\theta(z|x)$ is the prior distribution. $P_\theta(y|x,z)$ is the answer likelihood.

Efficient reasoning aims to learn the optimal policy $\theta^*$. It ensures the derivation of the correct answer $y^*$ while keeping the reasoning path concise. Therefore, the comprehensive utility $S(z)$ of a reasoning path is defined as:
\begin{equation}
S(z)=P_\theta(y^*|x,z)\cdot \eta(z).
\label{eq:eq2}
\end{equation}

This combines the accuracy likelihood and the efficiency evaluation function $\eta(z)$. The value of $\eta(z)$ decreases as the path length increases. The core optimization objective is to maximize the expected utility under the prior distribution:
\begin{equation}
\theta^*=\arg\max_\theta \mathbb{E}_{z\sim\pi_\theta(\cdot|x)}[S(z)].
\label{eq:eq3}
\end{equation}

However, directly optimizing Equation ~\ref{eq:eq3} given only the problem $x$ suffers from severe reward sparsity. High-quality trajectories that are both accurate and efficient are rarely sampled, severely limiting training efficiency and effectiveness.

\subsection{Theoretical Motivation: Utility Advantage of the Posterior Distribution}
\label{sec:Theoretical Motivation}
To overcome sampling difficulties, we adopt the concept of explanatory reconstruction from cognitive science. When explaining answers, humans eliminate invalid trial-and-error steps. They reversely construct an efficient logic chain to reach the result \citep{hoffrage2000hindsight}. In probabilistic models, this mechanism corresponds to the posterior distribution conditioned on the reference answer $y^*$:
\begin{equation}
q(z) = P_\theta(z|x, y^*) = \frac{P_\theta(y^*|x, z) \cdot \pi_\theta(z|x)}{P_\theta(y^*|x)} \propto P_\theta(y^*|x, z) \cdot \pi_\theta(z|x).
\label{eq:eq4}
\end{equation}
To mathematically verify its superiority as a guiding signal, we propose and prove the following proposition:

Proposition 1: Assuming that under the prior distribution $\pi_\theta(z|x)$, the correctness likelihood $L(z)$ of a reasoning path is non-negatively correlated with its comprehensive utility score $S(z)$, such that $Cov_{\pi}(L, S) \ge 0$. It follows that the expected utility of reasoning paths generated by the reference-answer-guided posterior distribution $q(z)$ is no less than that generated by the prior distribution:
\begin{equation}
\mathbb{E}_{z\sim q(z)}[S(z)]\ge \mathbb{E}_{z\sim\pi_\theta(\cdot|x)}[S(z)].
\label{eq:eq5}
\end{equation}
Please refer to Appendix ~\ref{sec:Detailed Mathematical} for the detailed proof and the necessity analysis of this condition.

\subsection{Variational Inference Framework and Objective Transformation}
\label{sec:Variational Inference}

Reference answers are invisible during the inference stage. Thus, the posterior distribution cannot be used directly. The core challenge is how to guide the prior distribution during training. It must internalize and replicate the efficient inference advantage established in Proposition 1. Therefore, we formalize the core objective of efficient inference. It is to maximize the expected utility under the prior distribution. For easier optimization, we further transform this objective. It maximizes the logarithm of this expectation:
\begin{equation}
\log J(\theta)=\log\mathbb{E}_{z\sim\pi_\theta(\cdot|x)}[P_\theta(y^*|x,z)\cdot \eta(z)].
\label{eq:eq6}
\end{equation}

We use the VI framework as the optimization basis. Note that we do not perform Bayesian inference on the true posterior $\pi_\theta(z|x,y^*)$. Instead, we treat the posterior $q(z)$ as a sampleable auxiliary distribution. We use the mathematical structure of the variational lower bound to connect posterior sampling with prior optimization. The following lower bound can be derived: 
\begin{equation}
\log J(\theta)\ge \text{ELBO}.
\label{eq:eq7}
\end{equation}

This lower bound is derived based on the standard Jensen's inequality (see Appendix ~\ref{sec:Derivation of the ELBO}):
\begin{equation}
\text{ELBO}=\mathbb{E}_{z\sim q(\cdot|x,y^*)}[\log P_\theta(y^*|x,z)+\log \eta(z)]-D_{\text{KL}}(q(z)||\pi_\theta(z|x)).
\label{eq:eq8}
\end{equation}

Its structure reveals two key operations. First, sample efficient paths from the posterior. Second, transfer posterior capabilities to the prior. 

\section{Method}

Based on the ELBO structure above Section ~\ref{sec:Variational Inference}, this paper proposes an efficient reasoning framework VPG-EA. As shown in Figure ~\ref{fig:Schematic}, VPG-EA instantiates distributions through a dual-stream sampling architecture. It uses cross-view evaluation to calculate utility scores. Finally, it completes knowledge transfer through variational distillation. The specific training algorithm process is detailed in Appendix ~\ref{sec:Algorithmic Procedure}.

\subsection{Generation and Construction of Posterior and Prior Distributions}

\begin{figure}[htbp]
  \centering
  \includegraphics[width=1\linewidth]{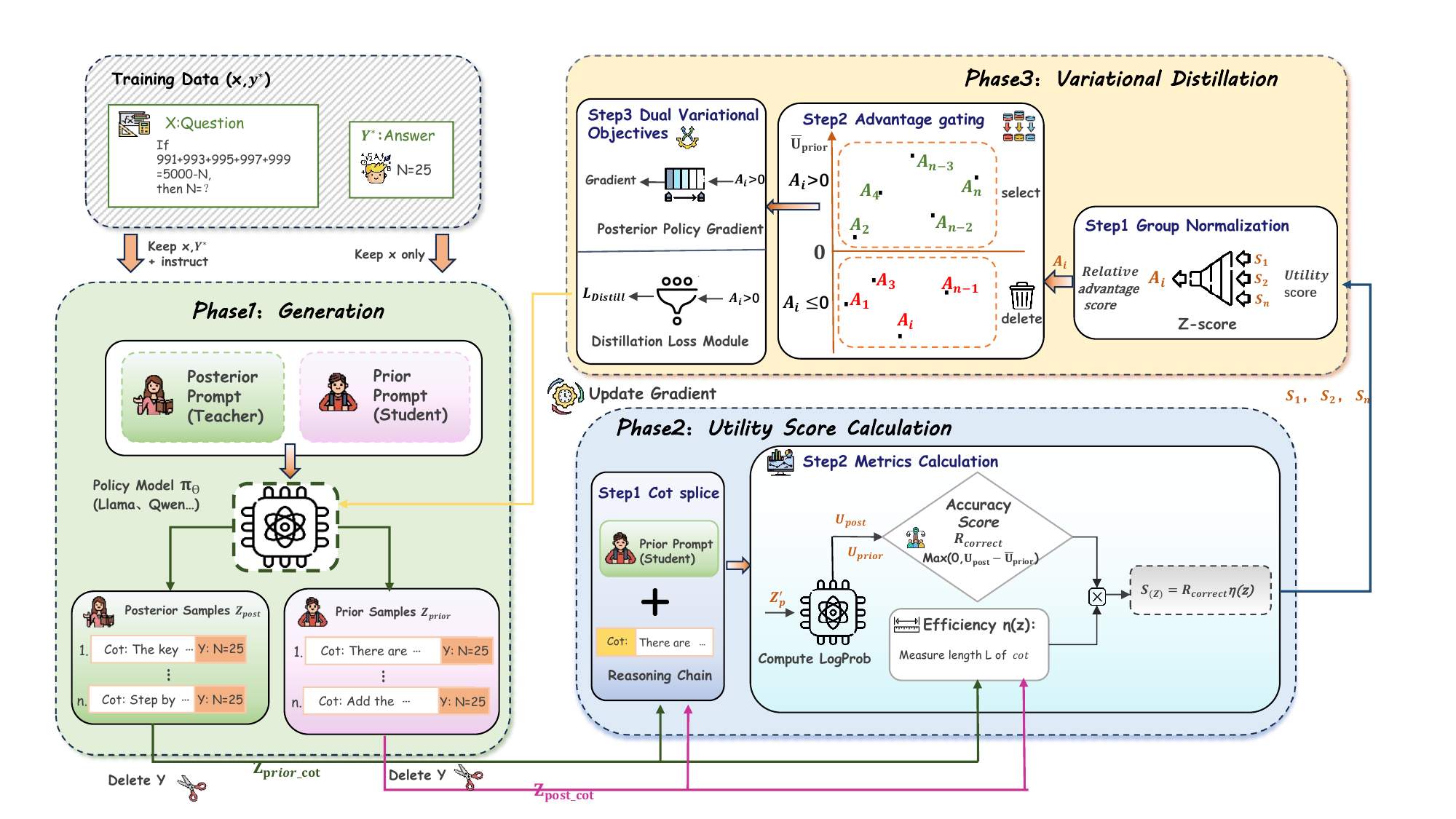}
  \caption{Overview of the VPG-EA framework, consisting of three phases: generation, utility score calculation, and variational distillation.}
  \label{fig:Schematic}
\end{figure}

Traditional VI requires two independent models to construct the prior and variational distributions. Training two models simultaneously in RL incurs unacceptable memory and computational overhead. To instantiate dual distributions within a single model, VPG-EA adopts the Parameter-shared Approximation of Conditional Distributions strategy. It uses differentiated system prompts to induce two distinct conditional distributions on the same set of parameters $\theta$. As shown in Figure ~\ref{fig:Schematic}, VPG-EA instantiates two parallel conditional distribution sampling streams on the same parameters $\theta$ using different System Prompts (see Appendix ~\ref{sec:Prompts} for prompt templates). These are the Teacher Stream (Posterior) and the Student Stream (Prior). The input of the teacher stream is $x \oplus y^* \oplus \text{Prompt}_{\text{teacher}}$. Since the reference answer $y^*$ is visible, this stream constructs an auxiliary distribution $q_\theta(z)$ conditioned on the answer. This serves as an engineering approximation of the variational posterior. Its generated reasoning path $z_{\text{post}}$ aims to explore the efficient reasoning manifold. The input of the student stream is $x \oplus \text{Prompt}_{\text{student}}$. This stream directly corresponds to the prior distribution $\pi_\theta(z|x)$. It represents the actual capability of the model during the inference phase.

\subsection{Cross-view Validation and Utility Score Calculation}

Standard ELBO assumes that samples from the posterior distribution naturally fit prior learning. However, during RL evaluation, the posterior distribution is conditioned on reference answers. Therefore, it may generate pseudo-efficient paths that the prior cannot reproduce. These paths score very high from the posterior view. But using them directly for prior optimization causes severe reward hacking and distribution shift. To ensure distribution consistency in knowledge transfer, we introduce cross-view validation. We consider a posterior path to contain transferable real logic only if it remains valid under prior conditions. Specifically, for a posterior path $z_{post}$, we explicitly strip the final generated answer from the trajectory. We evaluate its log-likelihood of deriving the correct answer under the prior distribution without answer guidance: $U_{post}=\log P_\theta(y^*|x,z_{post})$. Meanwhile, we calculate the average log-likelihood of prior paths in the same batch as a dynamic baseline $\bar{U}_{\text{prior}}$. This represents the current autonomous reasoning ability of the prior. We construct a distribution alignment filter:
\begin{equation}
R_{\text{correct}} = \max(0, U_{\text{post}} - \bar{U}_{\text{prior}}).
\label{eq:eq9}
\end{equation}

This truncation mechanism projects the posterior sample space to the region supported by the prior. It strictly filters out false high-scoring samples that rely on answer leakage. This ensures all paths entering the distillation phase have cross-distribution reproducibility. For validated samples, we further introduce relative length decay for efficiency modulation:
\begin{equation}
\eta(z) = \left( \frac{L_{\text{base}}}{L_{z_{\text{post}}}} \right)^\alpha,
\label{eq:eq10}
\end{equation}
here, $L_{base}$ is the average length of prior samples, $\alpha\ge0$ is a hyperparameter controlling efficiency sensitivity. Finally, the utility score of this posterior path $z_{post}$ is strictly calculated as the product of relative correctness and the efficiency coefficient:
\begin{equation}
\hat{S}(z_{post}) = R_{\text{correct}} \cdot \eta(z).
\label{eq:eq11}
\end{equation}
By scaling the log-advantage multiplicatively rather than using the additive $\log \eta(z)$ term from the formal ELBO, we ensure more stable efficiency modulation in RL. This score strictly decouples cross-distribution validation from efficiency constraints. It ensures that knowledge transfer in variational distillation is always anchored on the efficient manifold executable by the prior policy.

\subsection{Variational Distillation}

This section converts the theoretical ELBO structure in Equation ~\ref{eq:eq8} into a trainable RL objective. Directly using the absolute utility $\hat S_z$ as the advantage causes a severe cross-difficulty baseline deviation. Specifically, mediocre paths for simple problems often score higher than optimal paths for complex problems. This causes the model to over-converge on simple samples. Therefore, we apply Z-score normalization to the utilities of $G$ posterior paths sampled from the same group:
\begin{equation}
A_{z_i} = \frac{\hat S_{z_i} - \text{mean}(\hat S_{z_1},...,\hat S_{z_G})}{\text{std}(\hat S_{z_1},...,\hat S_{z_G}) + \epsilon}.
\label{eq:eq12}
\end{equation}
By obtaining the in-group relative advantage $A_{z_i}$, this operation removes the absolute difficulty scale of the problems. It accurately distinguishes the quality of the in-group samples. Second, inspired by the second term of ELBO, we instantiate knowledge transfer as an advantage-gated forward KL divergence Monte Carlo estimation:
\begin{equation}
L_{\text{Distill}} \approx \frac{1}{G} \sum_{i=1}^{G} [\mathbb{I}(A_{z_i} > 0) \cdot(sg[\log q_\theta(z_i|x, y^*)]-\log \pi_\theta(z_i|x))],
\label{eq:eq13}
\end{equation}
here, $\text{sg}[\cdot]$ denotes the stop-gradient operator ensuring unidirectional transfer, and the indicator function $\mathbb{I}(A_{z_i}>0)$ only retains high-quality samples with positive in-group advantages. This prevents low-quality paths from contaminating the prior policy. Combining the above components, the joint training objective of VPG-EA is:
\begin{equation}
L_{\text{Total}} \approx -\frac{1}{G} \sum_{i=1}^{G} [A_{z_i} \cdot \log q_{\theta}(z_i|x, y^*) ] + \beta L_{\text{Distill}},
\label{eq:eq14}
\end{equation}
here, $\beta=1$ serves as a heuristic initialization aligned with the ELBO structure. It is also the default parameter setting for the main experiments.
It should be noted that, limited by the discrete generation space, Equation ~\ref{eq:eq14} is not a strict mathematical implementation of ELBO. Instead, it is a heuristic instantiation using ELBO as a conceptual framework. The first term drives the posterior policy to continuously explore high-efficiency manifolds. The second term unidirectionally distills verified patterns to the prior policy through gating.

\section{Experiments}

\subsection{Setup}

\paragraph{Models and Baselines}
VPG-EA is instantiated using DeepSeek-R1-Distill-Qwen-1.5B and 7B \citep{guo2025deepseek} as the backbone models. Cross-architecture generalization is further verified with DeepSeek-R1-Distill-Llama-8B \citep{guo2025deepseek}. These base models are also adopted by multiple strong baselines varying in training paradigms and datasets: AdaptThink \citep{zhang2025adaptthink}, Laser \citep{liu2026learn}, AutoThink \citep{tu2025learning}, and JET \citep{han2026your} rely solely on RL; Thinkless \citep{fang2025thinkless} combines SFT with RL; and R1-VeriThinker \citep{chen2025verithinker} utilizes only SFT. Notably, AdaptThink, Laser, AutoThink, and Thinkless share the same training data source as the proposed method. Implementation details and baseline descriptions are provided in Appendix ~\ref{sec:Experimental Details}.

\paragraph{Datasets and Evaluation}
For training, the DeepScaleR dataset \citep{luo2025deepscaler} is employed, which contains approximately 40,000 high-quality math question-answer pairs with difficulty grading. For evaluation, a diverse mathematical reasoning test set with increasing difficulty is constructed, including GSM8K \citep{cobbe2021training}, MATH-500 \citep{hendrycks2021measuring}, AIME 2024 \footnote{\url{https://huggingface.co/datasets/math-ai/aime24}}, and AIME 2025 \footnote{\url{https://huggingface.co/datasets/math-ai/aime25 }}. To verify generalization beyond mathematics, two non-mathematical domain datasets are introduced: GPQA-Diamond \citep{DBLP:journals/corr/abs-2311-12022} and an MMLU-PRO subset \citep{wang2024mmlu} created by randomly sampling 50 questions from each non-mathematical discipline. Pass@1 \citep{chen2021evaluating} Accuracy (ACC) and Average Tokens (A.Tok) in the CoT are reported. To quantify the trade-off, the comprehensive metric $\varepsilon^3 = \frac{ACC^2}{A.Tok}$ \citep{han2026your} is adopted, which penalizes redundancy while prioritizing correctness.

\subsection{Comparative Experimental Results}
Table ~\ref{tab:comparative_results} demonstrates VPG-EA's significant and consistent advantages across both parameter scales. On the 1.5B and 7B models, VPG-EA achieves average comprehensive metrics $\varepsilon^3$ of 4.11 and 5.45, outperforming the best baselines by 8.73\% and 12.37\% and the base models by 69.14\% and 40.46\%, respectively. Notably, these performance margins widen as task difficulty and model parameter scale increase, with VPG-EA attaining maximum or near-maximum scores on 83\% of all metrics under identical training settings. This dominance persists across diverse training paradigms: while composite strategies show marginal gains, they significantly underperform VPG-EA, confirming that addressing the fundamental exploration bottleneck through a bottom-up RL paradigm yields far greater improvements than surface-level reward engineering or data-tuning combinations. Furthermore, VPG-EA consistently dominates across various training paradigms. It achieves state-of-the-art or competitive results across nearly all metrics when compared to baselines sharing the identical RL setting and datasets. More importantly, it significantly outperforms other composite strategies. This confirms that addressing the fundamental exploration bottleneck through a bottom-up RL paradigm yields far greater improvements in reasoning efficiency than simple data-tuning combinations or surface-level reward engineering.

\begin{table}[htb]
  \caption{In-domain comparison on mathematical reasoning benchmarks. $^\dagger$denotes methods trained with the same data and RL paradigm.}
  \label{tab:comparative_results}
  \centering
  \scriptsize 
  \setlength{\tabcolsep}{2.0pt}
  
  \begin{tabular}{l ccc ccc ccc ccc ccc}
    \toprule
    \multirow{2}{*}{Method} & \multicolumn{3}{c}{GSM8K} & \multicolumn{3}{c}{MATH-500} & \multicolumn{3}{c}{AIME24} & \multicolumn{3}{c}{AIME25} & \multicolumn{3}{c}{Average} \\
    \cmidrule(lr){2-4} \cmidrule(lr){5-7} \cmidrule(lr){8-10} \cmidrule(lr){11-13} \cmidrule(lr){14-16}
    & ACC$\uparrow$ & A.Tok$\downarrow$ & $\varepsilon^3\uparrow$ & ACC$\uparrow$ & A.Tok$\downarrow$ & $\varepsilon^3\uparrow$ & ACC$\uparrow$ & A.Tok$\downarrow$ & $\varepsilon^3\uparrow$ & ACC$\uparrow$ & A.Tok$\downarrow$ & $\varepsilon^3\uparrow$ & ACC$\uparrow$ & A.Tok$\downarrow$ & $\varepsilon^3\uparrow$ \\
    \midrule
    \multicolumn{16}{c}{\textbf{DeepSeek-R1-Distill-Qwen-1.5B}} \\
    \midrule
    Original Model & 77.02 & 756.52 & 7.84 & 75.00 & 3098.02 & 1.82 & 20.00 & 11524.67 & 0.03 & 16.67 & 12845.30 & 0.02 & 47.17 & 7056.13 & 2.43 \\
    Thinkless & \underline{83.60} & 645.84 & 10.82 & 81.20 & 1936.46 & \underline{3.40} & 20.00 & 7004.40 & 0.06 & 16.67 & 7661.17 & 0.04 & 50.37 & 4311.97 & 3.58 \\
    JET & 78.01 & \textbf{505.85} & \underline{12.03} & \textbf{83.80} & 2397.60 & 2.93 & \underline{26.67} & \textbf{6159.30} & \underline{0.12} & 16.67 & \textbf{5349.23} & 0.05 & 51.29 & \textbf{3602.99} & \underline{3.78} \\
    AdaptThink$^\dagger$ & 80.67 & 614.97 & 10.58 & 79.20 & 1960.20 & 3.20 & 23.33 & 7596.63 & 0.07 & \textbf{26.67} & 7365.07 & \textbf{0.10} & \underline{52.46} & 4384.22 & 3.49 \\
    Laser$^\dagger$ & 79.45 & 666.62 & 9.47 & 81.60 & 1911.50 & 3.20 & \underline{26.67} & 8457.90 & 0.08 & \underline{20.00} & 6695.57 & \underline{0.06} & 51.93 & 4432.90 & 3.20 \\
    AutoThink$^\dagger$ & \textbf{83.70} & 655.84 & 10.68 & 79.00 & \textbf{1845.33} & 3.38 & \underline{26.67} & \underline{6325.20} & 0.11 & \underline{20.00} & 7602.73 & 0.05 & 52.34 & 4107.28 & 3.56 \\
    VPG-EA$^\dagger$ & 81.00 & \underline{519.90} & \textbf{12.62} & \underline{82.40} & \underline{1891.80} & \textbf{3.59} & \textbf{33.33} & 6659.73 & \textbf{0.17} & \underline{20.00} & \underline{6678.70} & \underline{0.06} & \textbf{54.18} & \underline{3937.53} & \textbf{4.11} \\
    \midrule
    \multicolumn{16}{c}{\textbf{DeepSeek-R1-Distill-Qwen-7B}} \\
    \midrule
    Original Model & 89.90 & 700.90 & 11.53 & 91.00 & 2336.10 & 3.54 & \underline{53.33} & 8208.73 & 0.35 & \underline{33.33} & 10355.10 & 0.11 & \underline{66.88} & 5400.21 & 3.88 \\
    R1-VeriThinker & 71.50 & \textbf{367.40} & \underline{13.91} & 91.00 & 1726.20 & 4.80 & \underline{53.33} & 6127.47 & \underline{0.46} & \textbf{36.67} & \underline{6364.10} & \textbf{0.21} & 63.12 & \underline{3646.29} & \underline{4.85} \\
    JET & 87.87 & 558.62 & 13.82 & 88.20 & 1840.30 & 4.23 & 43.33 & 7181.60 & 0.26 & \underline{33.33} & 9850.33 & 0.12 & 63.18 & 4857.71 & 4.61 \\
    AdaptThink$^\dagger$ & 90.60 & 687.24 & 11.94 & 89.60 & 2028.86 & 3.96 & 43.33 & 8044.07 & 0.23 & \textbf{36.67} & 7998.13 & 0.17 & 65.05 & 4689.58 & 4.08 \\
    Laser$^\dagger$ & \underline{91.80} & 644.66 & 13.07 & \underline{92.60} & \underline{1705.36} & \underline{5.03} & 40.00 & \textbf{4877.27} & 0.33 & 26.67 & \textbf{5944.73} & 0.12 & 62.77 & \textbf{3293.01} & 4.64 \\
    AutoThink$^\dagger$ & 89.46 & 809.37 & 9.89 & 87.00 & 1945.81 & 3.89 & 40.00 & \underline{6125.50} & 0.26 & 30.00 & 6852.53 & 0.13 & 61.62 & 3933.30 & 3.54 \\
    VPG-EA$^\dagger$ & \textbf{92.00} & \underline{537.87} & \textbf{15.74} & \textbf{93.00} & \textbf{1626.01} & \textbf{5.32} & \textbf{56.67} & 6152.83 & \textbf{0.52} & \textbf{36.67} & 6762.60 & \underline{0.20} & \textbf{69.59} & 3769.83 & \textbf{5.45} \\
    \bottomrule
  \end{tabular}
\end{table}

\subsection{Generalization Experimental Results}

As can be seen from the comparative results of the non-mathematical domain generalization experiments shown in Table ~\ref{tab:Out-of-domain}, although VPG-EA does not achieve the optimum on the MMLU-Pro dataset at the 1.5B parameter scale, its performance approaches the SOTA level among the baselines. As the parameter scale expands to 7B, the performance ranking of the model on the two non-mathematical datasets significantly improves, with almost all metrics ranking first or second. The above results indicate that the VPG-EA framework possesses good universality: the increase in parameter scale effectively enhances its generalization ability. The model not only avoids catastrophic forgetting in non-mathematical disciplines, but conversely, by eliminating the cognitive redundancy brought about by overthinking, it enhances the direct answering capability in diverse scientific domains.

\begin{table}[htb]
  \caption{Out-of-domain generalization on GPQA-Diamond and MMLU-Pro. $^\dagger$denotes methods trained with the same data and RL paradigm.}
  \label{tab:Out-of-domain}
  \centering
  \scriptsize 
  \setlength{\tabcolsep}{4pt}
  
  \begin{tabular}{l ccc ccc ccc}
    \toprule
    \multirow{2}{*}{Method} & \multicolumn{3}{c}{GPQA-D} & \multicolumn{3}{c}{MMLU-Pro} & \multicolumn{3}{c}{Average} \\
    \cmidrule(lr){2-4} \cmidrule(lr){5-7} \cmidrule(lr){8-10}
    & ACC$\uparrow$ & A.Tok$\downarrow$ & $\varepsilon^3\uparrow$ & ACC$\uparrow$ & A.Tok$\downarrow$ & $\varepsilon^3\uparrow$ & ACC$\uparrow$ & A.Tok$\downarrow$ & $\varepsilon^3\uparrow$ \\
    \midrule
    \multicolumn{10}{c}{\textbf{DeepSeek-R1-Distill-Qwen-1.5B}} \\
    \midrule
    Original Model & 29.79 & 7419.90 & 0.12 & 24.76 & 4168.47 & 0.15 & 27.28 & 5794.19 & 0.14 \\
    Thinkless & \textbf{32.82} & 7465.22 & 0.14 & \textbf{31.08} & 3342.29 & \textbf{0.29} & \textbf{31.95} & 5403.76 & \textbf{0.22} \\
    JET & 26.77 & 6140.98 & 0.12 & 25.08 & \underline{2928.34} & \underline{0.21} & 25.93 & 4534.66 & 0.17 \\
    AdaptThink$^\dagger$ & 28.70 & \textbf{5106.90} & \textbf{0.16} & 23.23 & 3022.77 & 0.18 & 25.97 & \textbf{4064.84} & 0.17 \\
    Laser$^\dagger$ & \underline{30.80} & 6695.57 & 0.14 & \underline{25.38} & \textbf{2242.31} & \textbf{0.29} & \underline{28.09} & \underline{4468.94} & \textbf{0.22} \\
    AutoThink$^\dagger$ & 26.77 & \underline{5768.08} & 0.12 & 21.23 & 3544.66 & 0.13 & 23.97 & 4656.37 & 0.13 \\
    VPG-EA$^\dagger$ & \underline{30.80} & 6237.98 & \underline{0.15} & 25.23 & 3030.35 & \underline{0.21} & 28.02 & 4634.17 & \underline{0.18} \\
    \midrule
    \multicolumn{10}{c}{\textbf{DeepSeek-R1-Distill-Qwen-7B}} \\
    \midrule
    Original Model & 36.86 & 6471.26 & 0.21 & 47.53 & 4115.27 & 0.55 & 42.20 & 5293.27 & 0.38 \\
    R1-VeriThinker & 36.86 & 4060.43 & 0.33 & 43.85 & 4156.93 & 0.46 & 40.36 & 4108.68 & 0.40 \\
    JET & 36.89 & \textbf{2915.79} & \textbf{0.47} & \textbf{49.54} & 3916.35 & 0.63 & 43.22 & \textbf{3416.07} & \textbf{0.55} \\
    AdaptThink$^\dagger$ & \textbf{48.98} & 7124.44 & 0.34 & 48.46 & \underline{3696.78} & \underline{0.64} & \textbf{48.72} & 5410.61 & \underline{0.49} \\
    Laser$^\dagger$ & 33.83 & \underline{3603.66} & 0.32 & 43.07 & 4029.99 & 0.46 & 38.45 & 3816.83 & 0.39 \\
    AutoThink$^\dagger$ & 34.85 & 4763.04 & 0.25 & 44.77 & 4270.04 & 0.47 & 39.81 & 4516.54 & 0.36 \\
    VPG-EA$^\dagger$ & \underline{38.88} & 3892.62 & \underline{0.39} & \underline{48.77} & \textbf{3342.44} & \textbf{0.71} & \underline{43.83} & \underline{3617.53} & \textbf{0.55} \\
    \bottomrule
  \end{tabular}
\end{table}

\subsection{Ablation and Hyperparameter Studies}
\begin{figure}[htbp]
  \centering
  \begin{minipage}{0.54\linewidth}
    \centering
    \includegraphics[width=\linewidth]{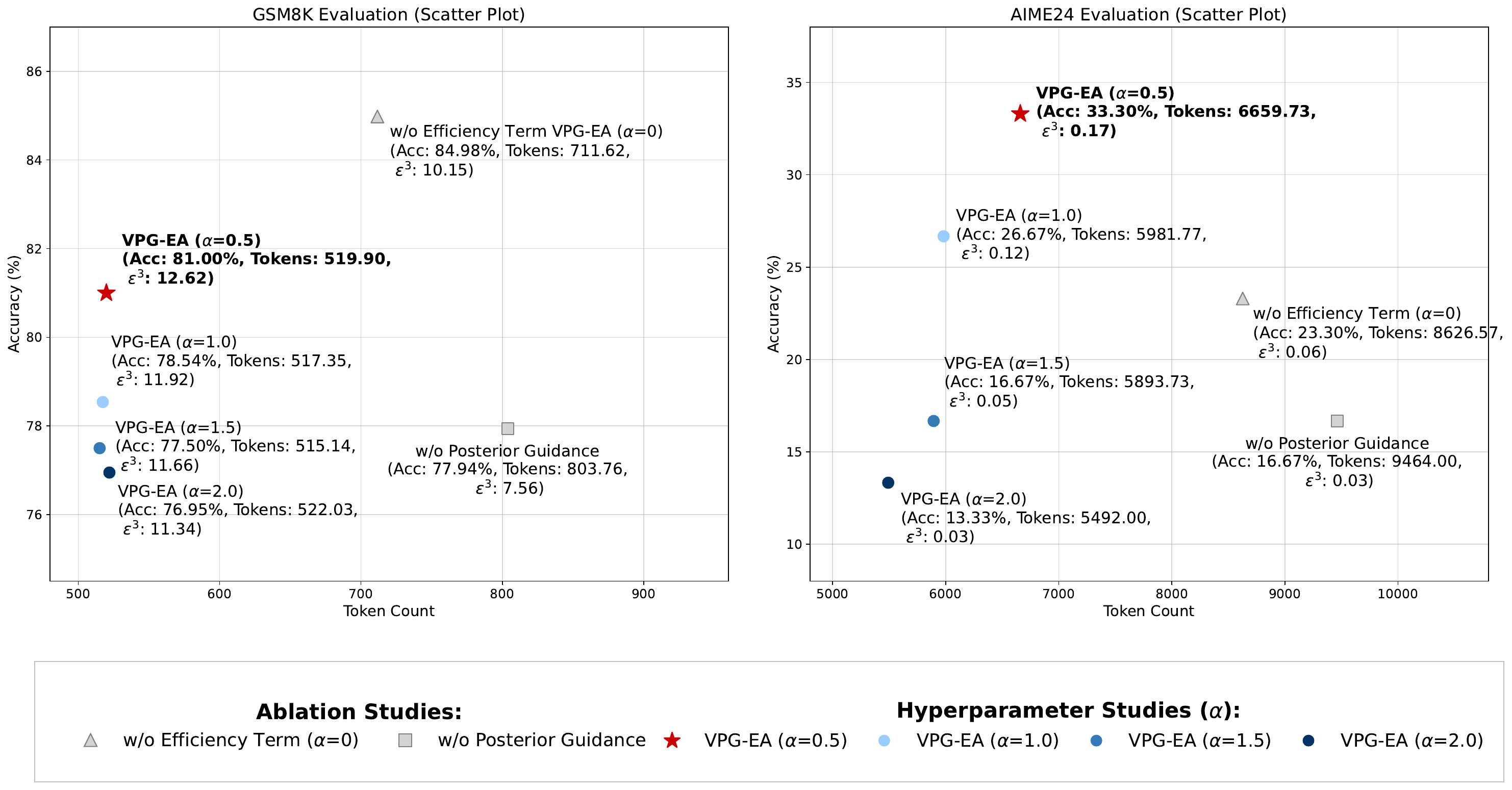}
    \caption{The performance of VPG-EA and ablation variants, as well as efficiency terms with different hyperparameters, on GSM8K and AIME24.}
  \label{fig:ablation_studies}
  \end{minipage}
  \hfill 
  \begin{minipage}{0.44\linewidth}
    \centering
    \includegraphics[width=\linewidth]{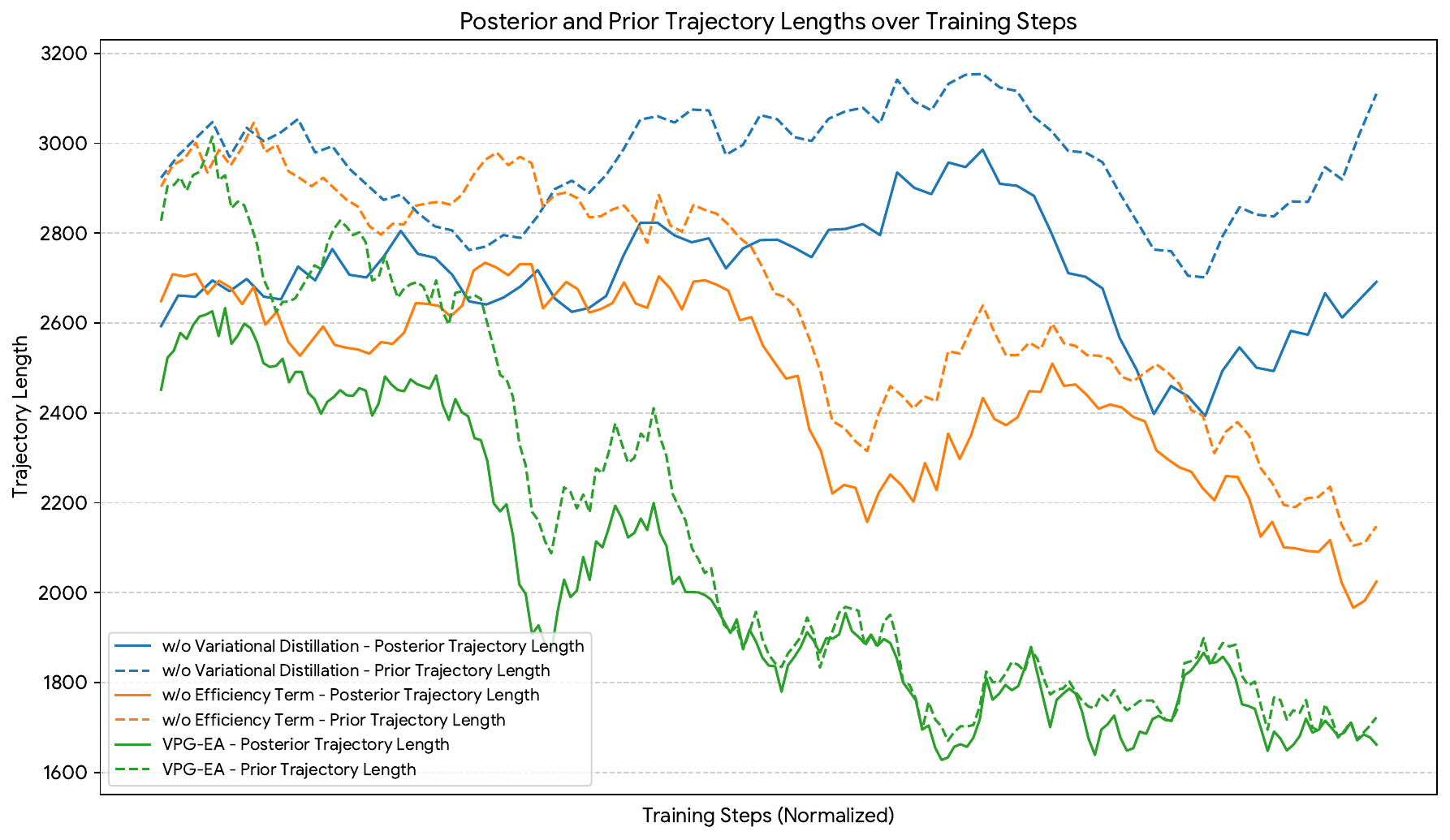}
    \caption{Comparison of training trajectory lengths between VPG-EA and its ablation variants (1.5B model).}
    \label{fig:Trajectory}
  \end{minipage}
\end{figure}

Figure ~\ref{fig:ablation_studies} presents ablation studies on GSM8K and AIME24, confirming the synergistic necessity of both posterior guidance and the efficiency constraint. Removing posterior guidance ($\beta=0$) severely degrades accuracy and inflates token counts, as evidenced by the AIME24 accuracy dropping to 16.67\% alongside a token surge to 9464, proving its critical role in overcoming sparse sampling to locate the efficient manifold. Conversely, removing the efficiency constraint ($\alpha=0$) strips the model of length control, resulting in an A.Tok of 8626 on AIME24 alongside a 10\% accuracy drop, highlighting this term as the key driver for converging to conciseness. Furthermore, the sensitivity parameter $\alpha$ defines the accuracy-efficiency trade-off boundary. We find that $\alpha=0.5$ yields the Pareto optimal balance, whereas excessively large values ($\alpha\ge 1.0$) force the model to aggressively sacrifice accuracy for marginal token reductions. This is exemplified on AIME24: as $\alpha$ increases from 0.5 to 2.0, Acc plummets from 26.67\% to 13.33\%, while Tok merely decreases from 5981 to 5492, imposing a penalty particularly detrimental on high-difficulty tasks. Finally, control experiments detailed in Appendix ~\ref{sec:Exploration of the Endogenous} confirm that VPG-EA's efficiency stems from an endogenous cognitive upgrade rather than mere mechanical obedience to instruction prompts.

\section{Discussion and Analysis}

\subsection{Dynamic Analysis of Reasoning Trajectory Length and Theoretical Validation of Variational Distillation}
Figure ~\ref{fig:Trajectory} tracks the prior and posterior reasoning trajectory lengths during the 1.5B model's training to analyze each module's impact. Removing variational distillation prevents the prior trajectory from decreasing, indicating that without posterior guidance, the model cannot spontaneously converge to an efficient reasoning manifold. Conversely, retaining distillation but removing the efficiency constraint allows the prior trajectory to decrease and approach the posterior, but the convergence is slow and unstable. Under the complete VPG-EA framework, both trajectories drop sharply and jointly converge. As a bottom-up RL paradigm, VPG-EA utilizes variational distillation to provide directional guidance and overcome inefficient exploration, while the efficiency constraint accelerates convergence. Theoretically, the posterior trajectory is initially shorter than the prior, which aligns with the non-negative expected utility boundary condition (Appendix ~\ref{sec:Detailed Mathematical}) and verifies Proposition 1. In late training stages, the prior trajectory drops below the posterior. Rather than an anomaly, this confirms that the posterior policy has successfully internalized high-quality reasoning patterns into the prior policy, completing its theoretical guidance role and achieving the endogenous transfer of efficient reasoning capabilities.

\subsection{Trade-off Analysis Under Difficulty Grading}
\label{sec:Trade-off Analysis}
\begin{figure}[htbp]
    \centering
    \begin{minipage}{0.49\linewidth}
    \centering
    \includegraphics[width=\linewidth]{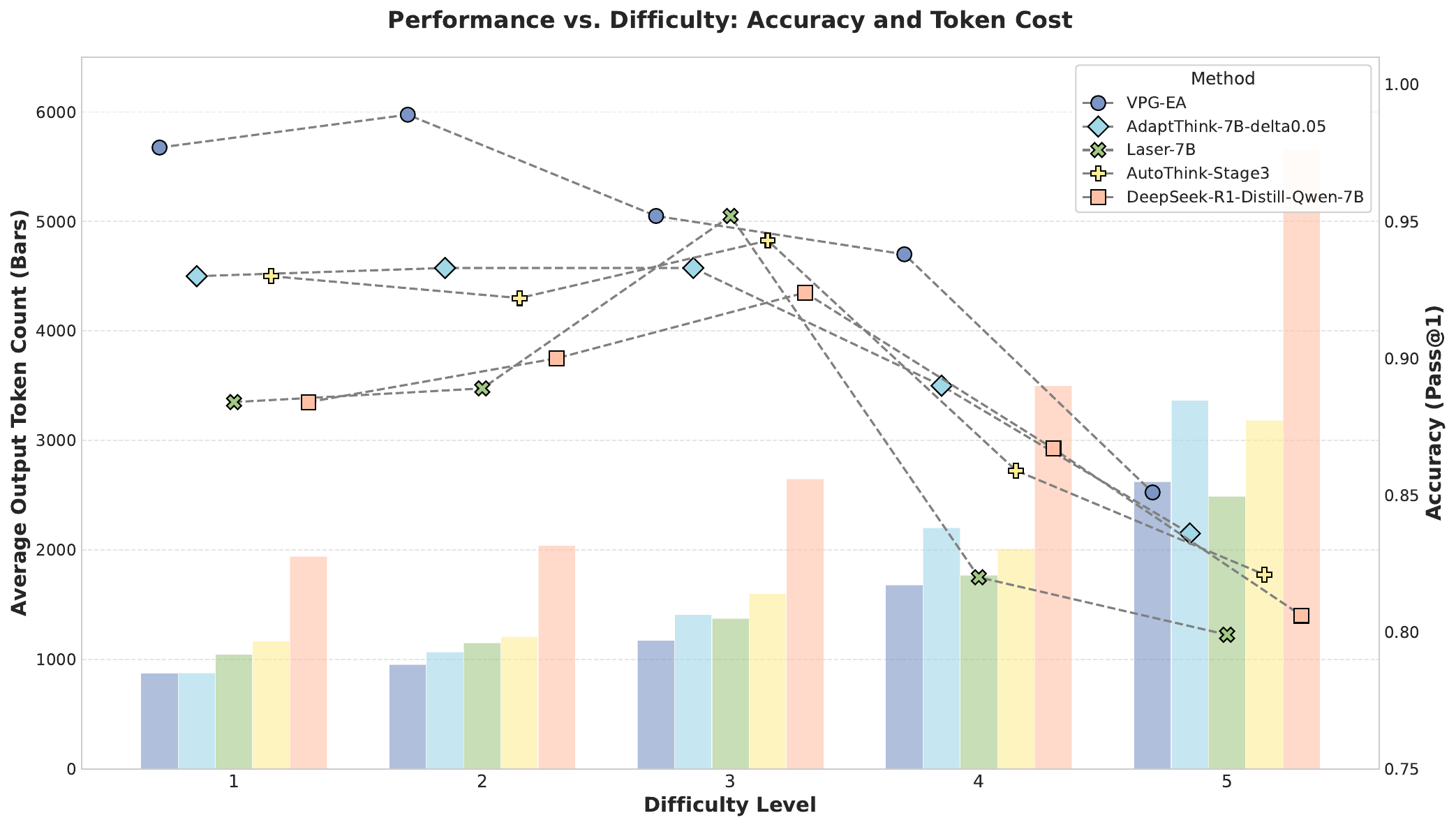}
    \caption{7B baseline performance across MATH-500 difficulty levels (1-5). Line: Accuracy; Bar: average output length.}
    \label{fig:performance_chart} 
    \end{minipage}
    \hfill
  \begin{minipage}{0.48\linewidth}
    \centering
    \includegraphics[width=\linewidth]{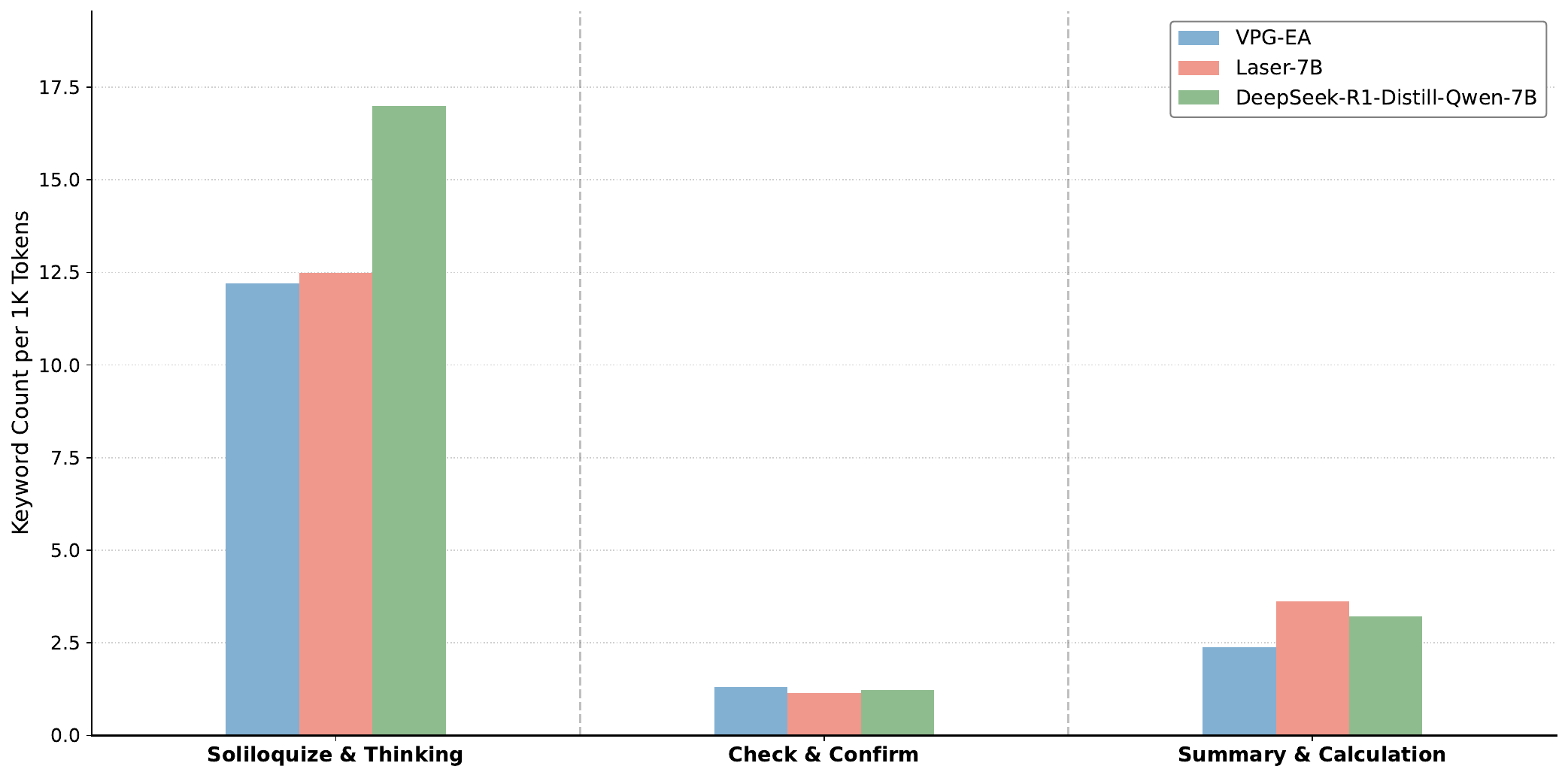}
    \caption{Frequency distribution of inference keywords per thousand tokens for each model under the MATH-500 benchmark.}
    \label{fig:keyword_chart}
  \end{minipage}
\end{figure}

Figure ~\ref{fig:performance_chart} evaluates model performance across the five difficulty levels of MATH-500. On levels 1 to 4, VPG-EA maintains high accuracy with shorter CoT outputs. At the highest difficulty, VPG-EA achieves a 5\% higher accuracy than Laser for a marginal cost of roughly 100 tokens, whereas Laser's accuracy degrades below that of the base model. Notably, other same-paradigm baselines suffer from both longer CoT lengths and lower accuracies on hard samples. This demonstrates that VPG-EA is highly robust to difficulty variations, successfully decoupling reasoning length from accuracy without sacrificing performance on complex problems.

\subsection{Micro-behavioral Analysis of Reasoning Trajectory}
\label{sec:Micro-Behavioral Analysis}
To decode how VPG-EA reshapes model behavior, we follow the taxonomy from \citep{hou2026thinkprune,DBLP:journals/tmlr/SuiCWZZYLWZZCH25} and categorize reasoning tokens into three groups using a rule-based keyword dictionary (see Appendix ~\ref{app:micro-behavior}): (i) Soliloquize \& Thinking, (ii) Check \& Confirm, and (iii) Summary \& Calculation. Figure ~\ref{fig:keyword_chart} reveals that while both VPG-EA and Laser effectively suppress tokens associated with the overthinking phenomenon (Soliloquize \& Thinking), their reallocation strategies diverge. Laser maintains logical coherence through verbose Summary \& Calculation steps. In contrast, VPG-EA aggressively compresses redundant summaries and selectively reallocates computational budget to the Check \& Confirm phase. This design aligns with established findings that explicit procedural verification is essential for rigorous reasoning \citep{weng2023large,dhuliawala2024chainofverification}. VPG-EA outperforms Laser because it strengthens logical verification at key nodes while compressing non-essential information. This enables VPG-EA to produce more rigorous outputs through leaner reasoning trajectories.

\section{Limitations}
\label{sec: Limitations}
Although the VPG-EA framework synergizes efficiency and accuracy, limitations remain. First, in RL, reward hacking is a fundamental paradigm problem for proxy objective functions \citep{wen2025language}. Even with cross-view evaluation and advantage gating, it is difficult to completely prevent models from exploiting topological loopholes in the policy space to find pseudo-logical shortcuts \citep{skalse2023invariance}. Second, the framework has a strong sampling dependency. The construction of the posterior distribution strictly depends on the mapping between the problem and the deterministic reference answer. Therefore, the framework currently only applies to verifiable tasks with clear evaluation criteria. Its applicability is limited in open-domain tasks that lack a single standard answer.

\section{Conclusion \& Future Work}
This paper addresses the reasoning inefficiency caused by overthinking in LLMs, revealing a fundamental limitation of traditional RL: blind exploration of the prior policy struggles to reach sparse, efficient manifolds that are both correct and concise. To overcome this, we introduce VI theory and propose the VPG-EA framework. Inspired by the ELBO structure, we construct a reference-answer-guided posterior distribution and derive an efficiency-aware variational objective for discrete reasoning spaces. Through dual-stream sampling, dynamic advantage gating, and variational distillation, VPG-EA effectively transfers efficient reasoning patterns to the prior policy. Cross-scale and cross-architecture experiments demonstrate that VPG-EA significantly reduces token consumption while maintaining or improving accuracy, offering a practical cost-reduction solution for industrial deployment. Future work will integrate VPG-EA with Process Reward Models for step-level verification, and extend the framework to open-domain tasks via multi-model consensus or human preference distributions as soft posterior guidance.

\bibliographystyle{plainnat} 
\bibliography{refs}


\appendix

\section{Detailed Mathematical Derivation and Theoretical Boundary Discussion of Proposition 1}
\label{sec:Detailed Mathematical}

In this appendix, we provide the complete proof of Proposition 1 (the expected utility advantage of the posterior) presented in Section~\ref{sec:Theoretical Motivation} of the main text. Our objective is to prove that, under the condition of a non-negative correlation between correctness and utility, the posterior distribution $q(z)$ conditioned on the reference answer $y^*$ generates reasoning paths with an expected utility that is strictly superior to that of the prior distribution $\pi_\theta(z|x)$.

\subsection{Notation Setup}
\label{sec:Notation Setup}

For clarity in derivation, we abbreviate the relevant variables as follows:

\begin{itemize}[leftmargin=*]
    \item Prior Distribution (Prior): $\pi(z) \triangleq \pi_\theta(z|x),$  
    
    \item Posterior Distribution (Posterior): $q(z) \triangleq \pi_\theta(z|x,y^*),$ 
    
    \item Correctness Likelihood: $L(z) \triangleq P(y^*|x,z),$ 

    \item Comprehensive Utility: $S(z) \triangleq L(z) \cdot \eta(z)$, where $\eta(z) > 0$ is the efficiency coefficient. 
\end{itemize}

According to Bayes' theorem, the relationship between the posterior distribution and the prior distribution is:
\begin{equation}
    q(z)=\frac{\pi(z)L(z)}{P(y^*|x)},
\end{equation}
where the denominator $P(y^*|x)$ is the marginal likelihood, representing the probability of the model answering correctly directly. It can be written in the form of an expectation under the prior:
\begin{equation}
    P(y^*|x)=\sum_{z \in Z} \pi(z)L(z)=E_{z \sim \pi}[L(z)].
\end{equation}

\subsection{Proof of the Accuracy Advantage of the Posterior Distribution over the Prior Distribution}

Before deriving the comprehensive utility advantage, we first mathematically prove an important prerequisite corollary: the expected likelihood of deriving the correct answer using the Bayesian posterior distribution constructed with the reference answer is necessarily greater than or equal to that of the prior distribution.

Expanding the expected correctness likelihood under the posterior distribution:
\begin{equation}
    E_{z \sim q(z)}[L(z)]=\sum_{z \in Z} q(z)L(z).
\end{equation}
Substituting the Bayesian expansion from Section~\ref{sec:Notation Setup} yields:
\begin{equation}
E_{z \sim q(z)}[L(z)]=\sum_{z \in Z} \left( \frac{\pi(z)L(z)}{E_{z \sim \pi}[L(z)]} \right) L(z)=\frac{\sum_z \pi(z)L^2(z)}{E_{z \sim \pi}[L(z)]}=\frac{E_{z \sim \pi}[L^2(z)]}{E_{z \sim \pi}[L(z)]}.
\end{equation}

To compare the difference in expected accuracy between the posterior and the prior, we subtract the two:
\begin{equation}
    E_{z \sim q(z)}[L(z)]-E_{z \sim \pi}[L(z)]=\frac{E_{z \sim \pi}[L^2(z)]-(E_{z \sim \pi}[L(z)])^2}{E_{z \sim \pi}[L(z)]}.
\end{equation}

According to the variance formula $Var(X)=E(X^2)-(E(X))^2$ in probability theory, the numerator of the above equation exactly corresponds to the variance of the correctness likelihood $L(z)$ under the prior distribution:
\begin{equation}
    E_{z \sim q(z)}[L(z)]-E_{z \sim \pi}[L(z)]=\frac{Var_{z \sim \pi}(L(z))}{E_{z \sim \pi}[L(z)]}.
\end{equation}

Since the variance of any random variable is strictly non-negative ($Var \ge 0$), and the denominator expectation $E_{z \sim \pi}[L(z)] > 0$ (assuming there exists a solution with a non-zero correctness probability in the prior space), we reach the following mathematical conclusion:
\begin{equation}
    E_{z \sim q(z)}[L(z)] \ge E_{z \sim \pi}[L(z)].
\end{equation}

This variance theorem establishes the inherent expected advantage of the true posterior distribution in terms of accuracy. However, the efficient reasoning objective of this paper requires not only correctness but also refinement of the reasoning process (introducing length decay $\eta(z)$). Therefore, whether the high expected correctness rate of the posterior can smoothly translate into the final high expected comprehensive utility $E_{z \sim q(z)}[S(z)]$ requires further covariance derivation.

\subsection{Covariance Transformation of Expected Utility}

We need to compare the magnitude relationship between the posterior expected utility $E_{q(z)}[S(z)]$ and the prior expected utility $E_\pi[S(z)]$.

Step 1: Expand the posterior expectation By the definition of expectation, the expected utility under the posterior distribution is:
\begin{equation}
    E_{z \sim q(z)}[S(z)]=\sum_{z \in Z} q(z)S(z)
\end{equation}

Step 2: Substitute the Bayesian formula Substituting $q(z)=\frac{\pi(z)L(z)}{P(y^*|x)}$ into the above equation:
\begin{equation}
    E_{z \sim q(z)}[S(z)]=\sum_{z \in Z} \left( \frac{\pi(z)L(z)}{P(y^*|x)} \right) S(z).
\end{equation}

Step 3: Extract the constant term and regroup Since $P(y^*|x)$ is a constant with respect to the summation variable $z$, it can be factored outside the summation sign. The numerator part becomes the weighted expectation under the prior distribution:
\begin{equation}
    E_{z \sim q(z)}[S(z)]=\frac{1}{P(y^*|x)}\sum_{z \in Z} \pi(z)(L(z) \cdot S(z))=\frac{E_{z \sim \pi}[L(z) \cdot S(z)]}{P(y^*|x)}.
\end{equation}

Step 4: Represent the marginal likelihood using the prior expectation Replacing the denominator $P(y^*|x)$ with $E_{z \sim \pi}[L(z)]$ (see Section~\ref{sec:Notation Setup}):
\begin{equation}
    E_{z \sim q(z)}[S(z)]=\frac{E_{z \sim \pi}[L(z) \cdot S(z)]}{E_{z \sim \pi}[L(z)]}.
\end{equation}

Step 5: Construct the inequality We need to prove $E_{q(z)}[S(z)] \ge E_\pi[S(z)]$. Substituting the result from Step 4 into the left side of the inequality, we must prove:
\begin{equation}
    \frac{E_{z \sim \pi}[L(z) \cdot S(z)]}{E_{z \sim \pi}[L(z)]} \ge E_{z \sim \pi}[S(z)].
\end{equation}

Step 6: Transform into covariance form Since the likelihood probability $L(z) \ge 0$, its expectation $E_\pi[L(z)]$ is a positive number. Multiplying both sides of the inequality by $E_\pi[L(z)]$, the direction of the inequality remains unchanged:
\begin{equation}
    E_{z \sim \pi}[L(z) \cdot S(z)] \ge E_{z \sim \pi}[L(z)] \cdot E_{z \sim \pi}[S(z)].
\end{equation}
Moving the right term to the left side:
\begin{equation}
    E_{z \sim \pi}[L(z) \cdot S(z)]-E_{z \sim \pi}[L(z)] \cdot E_{z \sim \pi}[S(z)] \ge 0.
\end{equation}

Step 7: Apply the definition of covariance According to the definition of covariance in statistics: for any random variables $X$ and $Y$, $Cov(X,Y)=E[XY]-E[X]E[Y]$.
Observing the expression in Step 6, the left side is exactly the covariance of $L(z)$ and $S(z)$ under the prior distribution $\pi$:
\begin{equation}
    Cov_{z \sim \pi}(L(z),S(z)) \ge 0.
\end{equation}

\subsection{Conclusion}
Through the above derivation, we conclude: the necessary and sufficient condition for Proposition 1 to hold is that under the prior distribution $\pi_\theta(z|x)$, the correctness likelihood $L(z)$ of the reasoning path is non-negatively correlated with the comprehensive utility $S(z)$, namely $Cov_{z \sim \pi}(L(z),S(z)) \ge 0$.

It should be clarified that because the comprehensive utility $S(z)=L(z) \cdot \eta(z)$ introduces a length-decreasing efficiency term $\eta(z)$, this non-negative covariance condition is not absolutely guaranteed to hold under all parameter settings. In the high-quality mathematical reasoning tasks established in this paper, when the efficiency coefficient $\eta(z)$ is subject to reasonable constraints—such as a moderate decay hyperparameter—the likelihood gain brought by the model reaching the correct logical manifold will remain dominant. Under such circumstances, a correct but slightly lengthy high-quality sample will still have a utility $S(z)$ that strictly dominates a vast number of erroneous and brief invalid exploration samples. Within this boundary, $L(z)$ and $S(z)$ maintain a positive correlation, and the advantage of posterior guidance strictly holds mathematically. When the framework adopts extreme hard length truncation, or if the efficiency decay coefficient is excessively large, even if the model derives the correct answer (an increase in $L(z)$), the length cost it incurs will cause a precipitous drop in $\eta(z)$, resulting in a comprehensive utility $S(z)$ that is actually lower than that of an incorrect but brief path. At this point, $L(z)$ and $S(z)$ will exhibit a negative correlation, and the posterior distribution will lose its expected utility advantage.

\section{Derivation of the ELBO}
\label{sec:Derivation of the ELBO}

In this appendix, we provide a detailed derivation of the ELBO proposed in Section~\ref{sec:Variational Inference} of the main text.

\paragraph{Original Objective}
Our objective is to maximize the logarithmic expected utility under the prior distribution $\pi_{\theta}(z|x)$: 
\begin{equation}
    \log J(\theta) = \log \mathbb{E}_{z \sim \pi_{\theta}(\cdot|x)} [S(z)].
\end{equation}
where the comprehensive utility is $S(z) = P_{\theta}(y^*|x, z) \cdot \eta(z)$. 

\paragraph{Importance Sampling Transformation}
Because it is exceedingly difficult for the prior distribution $\pi_{\theta}$ to sample high-reward regions during the early stages of training, directly optimizing the aforementioned objective is highly inefficient. We introduce the variational posterior distribution $q_{\theta}(z|x, y^*)$ and utilize the importance sampling identity to transform the objective function: 

Step 1: Expectation Expansion According to the definition of expectation, the objective function is expanded into a summation form over the discrete space $\mathcal{Z}$:
\begin{equation}
    \log J(\theta) = \log \sum_{z \in \mathcal{Z}} \pi_{\theta}(z|x) \cdot S(z).
\end{equation}

Step 2: Identity Insertion and Importance Sampling Multiply and divide by the posterior distribution $\pi_{\theta}(z|x, y^*)$ (abbreviated as $q(z)$) inside the summation: 
\begin{equation}
    \log J(\theta) = \log \sum_{z \in \mathcal{Z}} q(z) \frac{\pi_{\theta}(z|x)}{q(z)} \cdot S(z).
\end{equation}
At this point, the sampling distribution has shifted from the prior $\pi_{\theta}$ to the posterior $q(z)$, which can be rewritten as an expectation under the posterior: 
\begin{equation}
    \log J(\theta) = \log \mathbb{E}_{z \sim q(z)} \left[ \frac{\pi_{\theta}(z|x)}{q(z)} \cdot S(z) \right].
\end{equation}

Step 3: Applying Jensen's Inequality Since the logarithmic function $\log(\cdot)$ is a strictly concave function, according to Jensen's inequality $\log(\mathbb{E}[X]) \ge \mathbb{E}[\log(X)]$, we can move the logarithmic operation inside the expectation, thereby obtaining the lower bound of the objective function (ELBO): 
\begin{equation}
    \log J(\theta) \ge \mathbb{E}_{z \sim q(z)} \left[ \log \left( \frac{\pi_{\theta}(z|x)}{q(z)} \cdot S(z) \right) \right].
\end{equation}

Step 4: Log Expansion \& Rearrangement Using the logarithmic property $\log(abc/d) = \log a + \log b + \log c - \log d$, and expanding $S(z)$ into $P_{\theta}(y^*|x, z) \cdot \eta(z)$ :
\begin{align}
\text{ELBO} &= \mathbb{E}_{z \sim q(z)} \left[ \log \frac{P_{\theta}(y^*|x, z) \cdot \eta(z) \cdot \pi_{\theta}(z|x)}{q(z)} \right] \\
&= \mathbb{E}_{z \sim q(z)} \left[ \log P_{\theta}(y^*|x, z) + \log \eta(z) + \log \pi_{\theta}(z|x) - \log q(z) \right] \\
&= \mathbb{E}_{z \sim q(z)} \left[ \log P_{\theta}(y^*|x, z) + \log \eta(z) \right] - \mathbb{E}_{z \sim q(z)} \left[ \log q(z) - \log \pi_{\theta}(z|x) \right].
\end{align}

Step 5: Final Formula Utilizing the definition of KL divergence $D_{\text{KL}}(q || \pi) = \mathbb{E}_{q}[\log q - \log \pi]$, we obtain the final optimization objective: 
\begin{equation}
    \text{ELBO} = \mathbb{E}_{z \sim q(\cdot|x, y^*)} \left[ \log P_{\theta}(y^*|x, z) + \log \eta(z) \right] - D_{\text{KL}}(q(z) || \pi_{\theta}(z|x)).
\end{equation}

\section{Algorithmic Procedure}
\label{sec:Algorithmic Procedure}

\begin{algorithm}[htbp]
\caption{VPG-EA Training Procedure}
\label{alg:vpg_ea}
\begin{algorithmic}[1] 
\State \textbf{Input:} Dataset $\mathcal{D} = \{(x, y^*)\}$, Initial Policy $\pi_\theta$, KL coefficient $\beta$, Generation batch size $G$
\State \textbf{Output:} Optimized policy parameters $\theta^*$
\State Initialize policy parameters $\theta$
\While{not converged}
    \State Sample a batch of question-answer pairs $(x, y^*) \sim \mathcal{D}$
    
    \Statex \textit{\quad // Phase 1: Hybrid Generation}
    \State Sample $G$ prior paths: $\mathcal{Z}_{\text{prior}} = \{z_{\text{prior}}^{(1)}, \dots, z_{\text{prior}}^{(G)}\} \sim \pi_\theta(\cdot \mid x)$
    \State Sample $G$ posterior paths: $\mathcal{Z}_{\text{post}} = \{z_{\text{post}}^{(1)}, \dots, z_{\text{post}}^{(G)}\} \sim q_\theta(\cdot \mid x, y^*)$
    \State Compute average prior length: $L_{base} \leftarrow \frac{1}{G}\sum_{j=1}^{G} |z_{prior}^{(j)}|$
    
    \Statex \textit{\quad // Phase 2: Cross-Perspective Utility \& Advantage Estimation}
    \State Compute prior baseline: $\bar{U}_{\text{prior}} \gets \frac{1}{G} \sum_{j=1}^G \log P_\theta(y^* \mid x, z_{\text{prior}}^{(j)})$
    \For{$i = 1, \dots, G$}
        \State $U_{\text{post}}^{(i)} \gets \log P_\theta(y^* \mid x, z_{\text{post}}^{(i)})$
        \State $R_{\text{correct}}^{(i)} \gets \max\left(0, U_{\text{post}}^{(i)} - \bar{U}_{\text{prior}}\right)$
        \State $\eta_i \gets \Clamp\left( \left( \frac{L_{\text{base}}}{L_{z_{\text{post}}^{(i)}}} \right)^\alpha, \eta_{\min}, \eta_{\max} \right)$
        \State $\hat S_i \gets R_{\text{correct}}^{(i)} \cdot \eta_i$
    \EndFor
    \State Compute relative advantages $A_i \gets \frac{\hat S_i - \mu_s}{\sigma_s + \epsilon}$ for $i \in \{1, \dots, G\}$
    
    \Statex \textit{\quad // Phase 3: Variational Gradient \& Advantage-Gated Distillation}
    \State $\mathcal{L}_{\text{post\_pg}} \gets -\frac{1}{G} \sum_{i=1}^G A_i \log q_\theta(z_{\text{post}}^{(i)} \mid x, y^*)$
    \State $\mathcal{L}_{\text{distill}} \gets \frac{1}{G} \sum_{i=1}^G \mathbb{I}(A_i > 0) \left[ \sg (\log q_\theta(z_{\text{post}}^{(i)} \mid x, y^*))- \log \pi_\theta(z_{\text{post}}^{(i)} \mid x)\right]$
    \State $\mathcal{L}_{\text{total}} \gets \mathcal{L}_{\text{post\_pg}} + \beta \mathcal{L}_{\text{distill}}$
    \State Update parameters: $\theta \gets \theta - \gamma \nabla_\theta \mathcal{L}_{\text{total}}$
\EndWhile
\State \Return $\theta^*$
\end{algorithmic}
\end{algorithm}

As outlined in Algorithm ~\ref{alg:vpg_ea}, to align the theoretical derivation with practical RL training, we introduce the following key processes in our engineering implementation:

First, to ensure numerical stability during training, we apply a clamp operation to the efficiency coefficient $\eta(z)$. In practical sampling, the model occasionally generates abnormally short outputs, which causes a surge in the ratio $L_{base}/L_{z_{post}}$ and consequently triggers reward explosion. The introduction of the clamp mechanism effectively prevents such abnormal samples from dominating the gradient update, ensuring that the model does not deviate from the fundamental optimization of logical correctness while pursuing output conciseness. Specifically, in our main experiments, we explicitly set the bounds to $\eta_{min} = 0.5$ and $\eta_{max} = 2.0$ to balance efficiency incentives and training stability.

Second, during the variational distillation phase, to guarantee the strictly unidirectional transfer of knowledge, we apply a stop gradient (sg) operation to the posterior distribution $q(z)$. Because the prior and posterior streams share the identical set of model parameters, this operation restricts the gradient flow exclusively to the prior distribution $\pi_\theta$, effectively preventing the posterior distribution from degrading toward the prior distribution during the training process.

Finally, regarding the coefficient $\beta$ of the variational distillation term, we strictly adhere to the theoretical derivation of the ELBO presented in Section~\ref{sec:Variational Inference}, setting its value to 1 during the primary training phase. This parameter is explicitly retained in the algorithm procedure primarily to serve as a control variable for ablation studies (setting $\beta=0$ in the ablation variant where posterior guidance is removed).

\section{Prompts}
\label{sec:Prompts}

\subsection{Implementation Details of VPG-EA Prompts}

\begin{promptbox}[title={Student Prompt (Prior)}]
<|im\_start|>system\\
A conversation between user and assistant. The user asks a question, and the assistant solves it. The assistant first thinks about the reasoning process in the mind and then provides the user with the answer. The reasoning process is enclosed within <think></think> tags, i.e., <think>This is my reasoning.</think> This is my answer.<|im\_end|>

<|im\_start|>user\\
\{Question\}<|im\_end|>

<|im\_start|>assistant
\end{promptbox}

\begin{promptbox}[title={Teacher Prompt (Posterior)}]
<|im\_start|>system\\
A conversation between user and assistant. The user asks a question, and the assistant solves it. The assistant first thinks about the reasoning process in the mind and then provides the user with the answer. The reasoning process is enclosed within <think></think> tags, i.e., <think>This is my reasoning.</think> This is my answer.<|im\_end|>

<|im\_start|>user\\
Given the following question and its reference answer, your task is to produce a step-by-step explanation that logically leads to the reference answer, written in the style of a first-person think-aloud monologue. You are encouraged to draw on the reference answer for internal guidance to help structure and support your reasoning, but the final monologue must read as a genuine, first-encounter, real-time discovery, without mentioning or implying any prior access to the reference answer.\\
Question: \{Question\}\\
Reference Answer: \{Reference Answer\}
1. Output only the first-person, think-aloud monologue. Do not include any preface, summary, or restatement of these instructions.\\
2. Maintain the tone of a focused individual thinking to themself. Avoid meta-commentary like "for the first time," and any phrasing that reveals simulation.\\
3. Do not mention, imply, or hint at prior access to the Reference Answer in the monologue. Avoid phrases like "according to the answer..." or "to get to that answer...", and any euphemism that signals foreknowledge.\\
4. Do not merely restate the final answer in the monologue; articulate the reasoning pathway with sufficient intermediate steps, rationale, decision points, verification, and any necessary error-correction or backtracking.<|im\_end|>
<|im\_start|>assistant
\end{promptbox}

\subsection{Exploration of the Endogenous Attribution of Efficiency after Removing Instruction Bias}
\label{sec:Exploration of the Endogenous}

\begin{promptbox}[title={Student Prompt (Prior)}]
<|im\_start|>system\\
You are a student solving a problem. You must show your work.\\
First, provide your step-by-step reasoning process enclosed in <think></think> tags.\\
Second, provide your final answer enclosed in \textbackslash boxed\{\}.\\
<|im\_end|>

<|im\_start|>user\\
\{Question\}\\
<|im\_end|>

<|im\_start|>assistant
\end{promptbox}

\begin{promptbox}[title={Teacher Prompt (Posterior)}]
<|im\_start|>system\\
You are an expert teacher providing a model solution. You are given a question and its correct answer.\\
Your task is to generate a \textbf{concise, direct, and efficient} step-by-step solution that logically leads \textit{directly} to the given answer.\\
\textbf{Avoid} any unnecessary exploration, self-correction, or reflective 'think-aloud' monologues. Focus \textit{only} on the core, necessary logical steps.\\
The reasoning process must be enclosed in <think></think> tags. The final answer must be enclosed in \textbackslash boxed\{\}.\\
<|im\_end|>

<|im\_start|>user\\
Question: \{Question\}\\
Answer: \{Answer\}\\
<|im\_end|>

<|im\_start|>assistant
\end{promptbox}

When evaluating the efficient reasoning performance of the VPG-EA framework across multiple benchmarks, a core question requires exploration: does the reduction in model output length originate from the RL algorithm successfully locating and learning sparse, highly efficient samples, or is it merely because the model is fitting a certain instruction bias that implies a reduction in output? To eliminate the interference caused by such instructions, we designed a set of extreme comparative prompt experiments to verify the true source of the model's efficiency after training with the VPG-EA framework. Our original instructions (see D.1) not only do not require the model to remain concise, but rather explicitly encourage trial-and-error and exploration. We require the model to elaborate on the reasoning path in detail, including sufficient intermediate steps, underlying principles, decision points, verification processes, and any necessary error-correction or backtracking. As a baseline for the comparative prompts (see D.2), we constructed an extremely rigid efficiency instruction, requiring the model to generate a concise, direct, and efficient step-by-step solution, avoiding any unnecessary exploration, self-correction, or reflectiveness.

The experimental results, as shown in Table ~\ref{tab:performance_comparison}, indicate that the Instruction-Based Prompts achieve a shorter average output length on GSM8K and AIME24; however, this length advantage is accompanied by significant performance degradation: the accuracy on AIME24 plummets from 33.33\% to 26.67\%, and there are also varying degrees of decline on GSM8K and MATH-500. This demonstrates that while rigid length-constrained instructions can mechanically truncate the output, the cost is the destruction of the necessary computational depth required for the model to process complex reasoning tasks. In contrast, VPG-EA achieves a comprehensive lead in accuracy while maintaining a similar length level, proving that its length optimization is not a simple shortening, but a selective refinement. The aforementioned experiments powerfully prove that the effectiveness of VPG-EA does not rely on fitting efficient instructions. In our framework, even if the system instructions explicitly encourage divergence and trial-and-error, the model still achieves significant refinement in the final output. Therefore, the VPG-EA framework genuinely empowers the large model to find a logically rigorous and redundancy-free efficient manifold; this is a cognitive upgrade at the algorithmic level, rather than remaining at the level of superficial compliance to instructions.

\paragraph{Clarification on Evaluation Protocol and Data Leakage Prevention}
To address potential concerns regarding data integrity, we explicitly clarify the usage of the Answer field in the prompts. While the Posterior template in Appendix ~\ref{sec:Exploration of the Endogenous} includes a reference answer field to facilitate the construction of the posterior distribution during the training phase, this field was strictly omitted during the evaluation of all benchmarks (GSM8K, MATH-500, AIME, etc.). During inference, the model was provided only with the Question and the system instructions to ensure a zero-leakage, fair comparison with all baseline models. The performance gains observed in VPG-EA stem from the internalized efficient reasoning manifold learned during variational distillation, rather than any form of ground-truth injection during testing.

\begin{table}[htbp]
  \caption{Performance comparison between instruction-based efficiency prompts and the VPG-EA framework.}
  \label{tab:performance_comparison}
  \centering
  \scriptsize
  \setlength{\tabcolsep}{4pt}

  \begin{tabular}{l ccc ccc ccc}
    \toprule
    & \multicolumn{3}{c}{GSM8K} & \multicolumn{3}{c}{MATH-500} & \multicolumn{3}{c}{AIME24} \\
    \cmidrule(lr){2-4} \cmidrule(lr){5-7} \cmidrule(lr){8-10}
    Method & ACC $\uparrow$ & A.Tok $\downarrow$ & $\varepsilon^3$ $\uparrow$ & ACC $\uparrow$ & A.Tok $\downarrow$ & $\varepsilon^3$ $\uparrow$ & ACC $\uparrow$ & A.Tok $\downarrow$ & $\varepsilon^3$ $\uparrow$ \\
    \midrule
    VPG-EA & \textbf{81.00} & 519.90 & \textbf{12.62} & \textbf{82.40} & \textbf{1891.80} & \textbf{3.59} & \textbf{33.33} & 6659.73 & \textbf{0.17} \\
    Instruction-Based Prompts & 78.46 & \textbf{516.16} & 11.93 & 81.60 & 1965.03 & 3.39 & 26.67 & \textbf{5708.50} & 0.12 \\
    \bottomrule
  \end{tabular}
\end{table}

\section{Performance across Different Model Architectures}
\label{sec:Performance across}

To verify the cross-architecture generalization capability of VPG-EA, we conduct experiments on DeepSeek-R1-Distill-Llama-8B and compared it with advanced baselines at this parameter scale: Manifold Steering \citep{huang2025mitigating} By employing the mean difference method and PCA analysis, this approach extracts and purifies the low-dimensional manifold direction representing overthinking from the model's activation space, and subsequently applies ablation interventions to the residual stream activations at each layer, thereby mitigating redundant reasoning internally within the model. ACPO \citep{cheng2025incentivizing} This method introduces explicit system-aware tokens (e.g., \textless fast\_think\textgreater) for SFT cold start, and subsequently integrates an online token length budget mechanism and a composite reward function within the GRPO framework, enabling the model to achieve dynamic switching between fast and slow thinking modes and adaptive allocation of cognitive resources based on task difficulty.
The results are shown in Table~\ref{tab:cross-architecture}. Except for Base and VPG-EA, which are locally evaluated, the data for the remaining baselines are cited from their original papers.

The experimental results indicate that VPG-EA also achieves a good balance between efficiency and accuracy on Llama-8B. On GSM8K, VPG-EA improves the accuracy of the base model by 1.29\%, and the average token consumption decreases by 199.18, representing a reduction of 22.71\%; although Manifold Steering has lower token consumption (542), its accuracy significantly degrades. On MATH-500, VPG-EA optimizes the average token consumption from 3292.8 to 2536.2, with a simultaneous improvement in accuracy, yielding an overall performance superior to ACPO and comparable to R1-VeriThinker-8B. On AIME24, the base model consumes 12588.47 tokens to achieve only 33.33\% accuracy, whereas VPG-EA increases the accuracy to 40\% and reduces Token consumption by approximately 3000. Although some baselines report higher accuracy, VPG-EA achieves better computational efficiency while ensuring relatively high accuracy, indicating that its core mechanism can guide the model to autonomously internalize the efficient reasoning manifold, realizing a dual improvement in correctness and efficiency without relying on specific architectures.

\begin{table}[htbp]
  \caption{Cross-architecture generalization experimental results of VPG-EA.}
  \label{tab:cross-architecture}
  \centering
  \scriptsize 
  \setlength{\tabcolsep}{4pt} 
  
  \begin{tabular}{l ccc ccc ccc ccc}
    \toprule
    \multirow{2}{*}{Model} & \multicolumn{3}{c}{GSM8K} & \multicolumn{3}{c}{MATH-500} & \multicolumn{3}{c}{AIME24} & \multicolumn{3}{c}{Average} \\
    \cmidrule(lr){2-4} \cmidrule(lr){5-7} \cmidrule(lr){8-10} \cmidrule(lr){11-13}
    & ACC$\uparrow$ & A.Tok$\downarrow$ & $\varepsilon^3\uparrow$ & ACC$\uparrow$ & A.Tok$\downarrow$ & $\varepsilon^3\uparrow$ & ACC$\uparrow$ & A.Tok$\downarrow$ & $\varepsilon^3\uparrow$ & ACC$\uparrow$ & A.Tok$\downarrow$ & $\varepsilon^3\uparrow$ \\
    \midrule
    Original Model & 87.87 & 877 & 8.80 & 87.40 & 3292.80 & 2.32 & 33.33 & 12977.53 & 0.09 & 69.53 & 5715.78 & 3.74 \\
    Manifold Steering* & 82.80 & \textbf{542} & \textbf{12.65} & 88.00 & 2873.00 & 2.70 & \textbf{50.00} & 9457.00 & \underline{0.26} & \underline{73.60} & \underline{4290.67} & \textbf{5.20} \\
    R1-VeriThinker* & - & - & - & \textbf{89.90} & 2953.00 & 2.74 & \underline{46.90} & 11285.00 & 0.19 & - & - & - \\
    ACPO* & - & - & - & 87.00 & \textbf{2232.00} & \textbf{3.39} & 43.30 & \textbf{7405.00} & 0.25 & - & - & - \\
    VPG-EA & \textbf{89.16} & \underline{677.82} & \underline{11.73} & \underline{89.60} & \underline{2536.20} & \underline{3.17} & 46.70 & 8225.07 & \textbf{0.27} & \textbf{75.14} & \textbf{3813.03} & \underline{5.05} \\
    \bottomrule
  \end{tabular}
  
  \vspace{6pt}
  \raggedright
  \textit{Note:} * denotes results cited directly from their original papers, as their model checkpoints are not publicly available. To ensure statistical fairness, average performance metrics are computed exclusively for models evaluated across all benchmarks. We omit cross-task aggregate comparisons for baselines with incomplete data.
\end{table}

\section{Experimental Details}
\label{sec:Experimental Details}

\paragraph{Hardware and Software Platforms}
The training infrastructure is based on a single-node configuration, equipped with four NVIDIA RTX Pro 6000 Blackwell (96GB) high-performance GPUs with a unified CUDA version of 12.4. The system runs on Ubuntu 22.04, utilizing a Python environment (version 3.12.3) and PyTorch 2.5.1. This study adopts the vLLM (0.7.3) framework for efficient batch decoding and employs the DeepSpeed framework during the training phase to optimize memory.

\paragraph{Training Configurations}
The experiment utilizes the AdamW optimizer, with the learning rate set to 1e-6 and a warmup ratio of 0.2. The global effective batch size is 64, and the sampling temperature is set to 1.0, paired with 4 prior and 4 posterior sampling examples respectively. Differentiated parallel strategies are applied based on model scales: the 1.5B model uses DeepSpeed ZeRO-1, while the 7B and 8B models enable DeepSpeed ZeRO-3 to optimize memory distribution. Additionally, the hyperparameters $\alpha$ and $\beta$ are fixed at 0.5 and 1.0, respectively, and the efficiency clamp bounds $[\eta_{min}, \eta_{max}]$ are set to $[0.5, 2.0]$ (as detailed in Appendix ~\ref{sec:Algorithmic Procedure}). Regarding the computational cost, a single complete training run for the 1.5B, 7B, and 8B models on 4 $\times$ NVIDIA RTX Pro 6000 GPUs takes approximately 8.77, 16.91, and 18.92 hours, respectively.

\paragraph{Evaluation Configurations}
The vLLM framework is adopted to accelerate inference, with the temperature set to 0.6, Top-p sampling set to 0.95, and the maximum output length set to 16K tokens.

\paragraph{baseline introduction}
\begin{itemize}[leftmargin=*]
    \item AdaptThink \citep{zhang2025adaptthink}: This method utilizes RL to endow the model with dynamic decision-making capabilities, enabling it to skip lengthy reasoning processes for simple problems and generate answers directly.
    
    \item Laser \citep{liu2026learn}: This method introduces a length-based reward shaping mechanism, dynamically penalizing excessively long chains of thought during RL training to guide the model to shorten the reasoning length while maintaining accuracy.
    
    \item AutoThink \citep{tu2025learning}: This method adopts a multi-stage RL framework, aiming to balance computational overhead and performance on problems of varying difficulty through adaptive reasoning strategies.

    \item JET \citep{han2026your}: Based on the DAPO RL framework, this method combines trajectory rollout generation and a progressive early-stopping truncation strategy to construct short reasoning trajectories, and designs a composite reward mechanism to drive training, allowing the model to actively terminate unnecessary reasoning processes.

    \item Thinkless \citep{fang2025thinkless}: This method constructs long-short response pairs through knowledge distillation and decouples the optimization process of reasoning mode selection and answer generation.

    \item R1-VeriThinker \citep{chen2025verithinker}: This method proposes a verification-based pruning strategy, utilizing an auxiliary verifier to identify and remove redundant steps in the reasoning path.
\end{itemize}

\section{Keyword Mapping for Micro-behavioral Analysis}
\label{app:micro-behavior}

The conceptual taxonomy of categorizing reasoning behavior into distinct cognitive phases is inspired by established frameworks in existing literature \citep{hou2026thinkprune,DBLP:journals/tmlr/SuiCWZZYLWZZCH25}. To operationalize this taxonomy for the micro-behavioral analysis presented in Section ~\ref{sec:Micro-Behavioral Analysis}, we developed a rule-based dictionary of explicit transitional phrases. These keywords were utilized to match and quantify the corresponding tokens generated within the \texttt{<think>...</think>} trajectories. The complete mapping between the behavior categories and their respective keywords is detailed in Table \ref{tab:keyword_mapping}. 

\begin{table}[htbp]
  \caption{Keyword mapping for reasoning behavior categorization.}
  \label{tab:keyword_mapping}
  \centering
  \small
  \begin{tabular}{p{0.3\linewidth} p{0.6\linewidth}}
    \toprule
    \textbf{Behavior Category} & \textbf{Keywords} \\
    \midrule
    \textbf{Soliloquize \& Thinking} & \texttt{Wait}, \texttt{But}, \texttt{wait}, \texttt{Hold on}, \texttt{Alternatively} \\
    \midrule
    \textbf{Check \& Confirm} & \texttt{Let me confirm}, \texttt{Double-check} \\
    \midrule
    \textbf{Summary \& Calculation}  & \texttt{Remember}, \texttt{Let me compute}, \texttt{Therefore} \\
    \bottomrule
  \end{tabular}
\end{table}

\section{Case Study}
\label{sec:Case Study}

To more intuitively demonstrate the practical effects of the VPG-EA framework in mitigating the overthinking problem in LLMs and achieving difficulty-adaptive computation allocation, this appendix provides specific qualitative analysis cases. We sample three mathematical reasoning questions with different difficulty gradients from the MATH-500 evaluation benchmark, and compare the microscopic behavioral differences in CoT generation between the base model DeepSeek-R1-Distill-Qwen-7B and the model trained with VPG-EA. 

As described in the quantitative analysis in Sections~\ref{sec:Trade-off Analysis} and ~\ref{sec:Micro-Behavioral Analysis} of the main text, the base model, when facing various types of problems, often generates a large amount of redundant monologue indicating hesitation and self-doubt, and even falls into the trap of sacrificing computational efficiency for marginal accuracy gains. Through the following case comparisons, it can be clearly observed that VPG-EA has successfully internalized the efficient logical manifold: on simple problems, it can discard invalid divergence and directly reach the answer; on complex problems, while significantly reducing meaningless trial-and-error, it accurately retains the procedural verification steps necessary to ensure correctness, thereby achieving an excellent balance between reasoning rigor and computational efficiency across all difficulty levels.

\begin{figure}[htbp]
  \centering
  \includegraphics[width=1\linewidth]{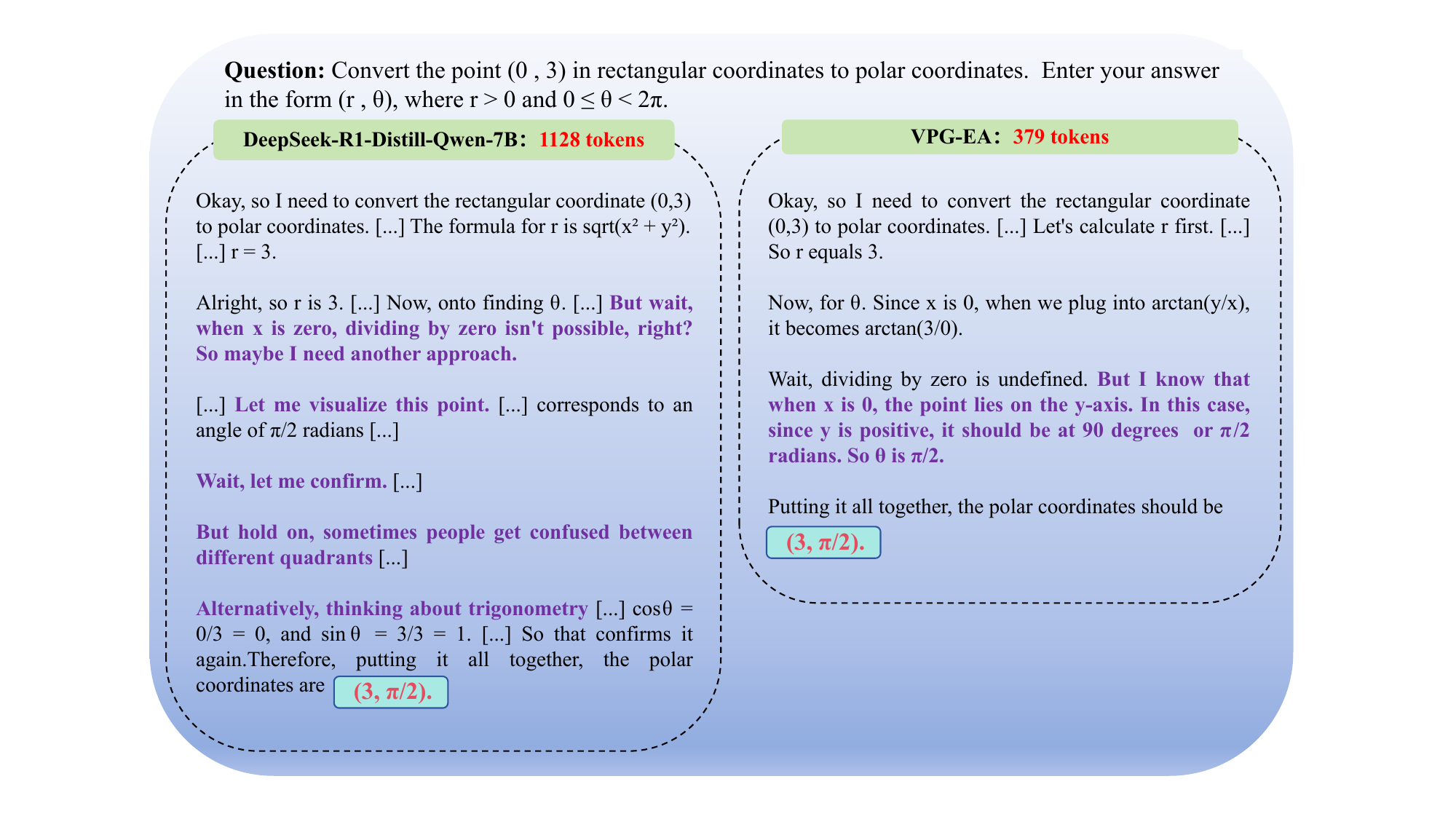}
  \caption{Comparison of reasoning cases for simple difficulty in MATH-500.}
  \label{fig:Case_study}
\end{figure}

\begin{figure}[htbp]
  \centering
  \includegraphics[width=1\linewidth]{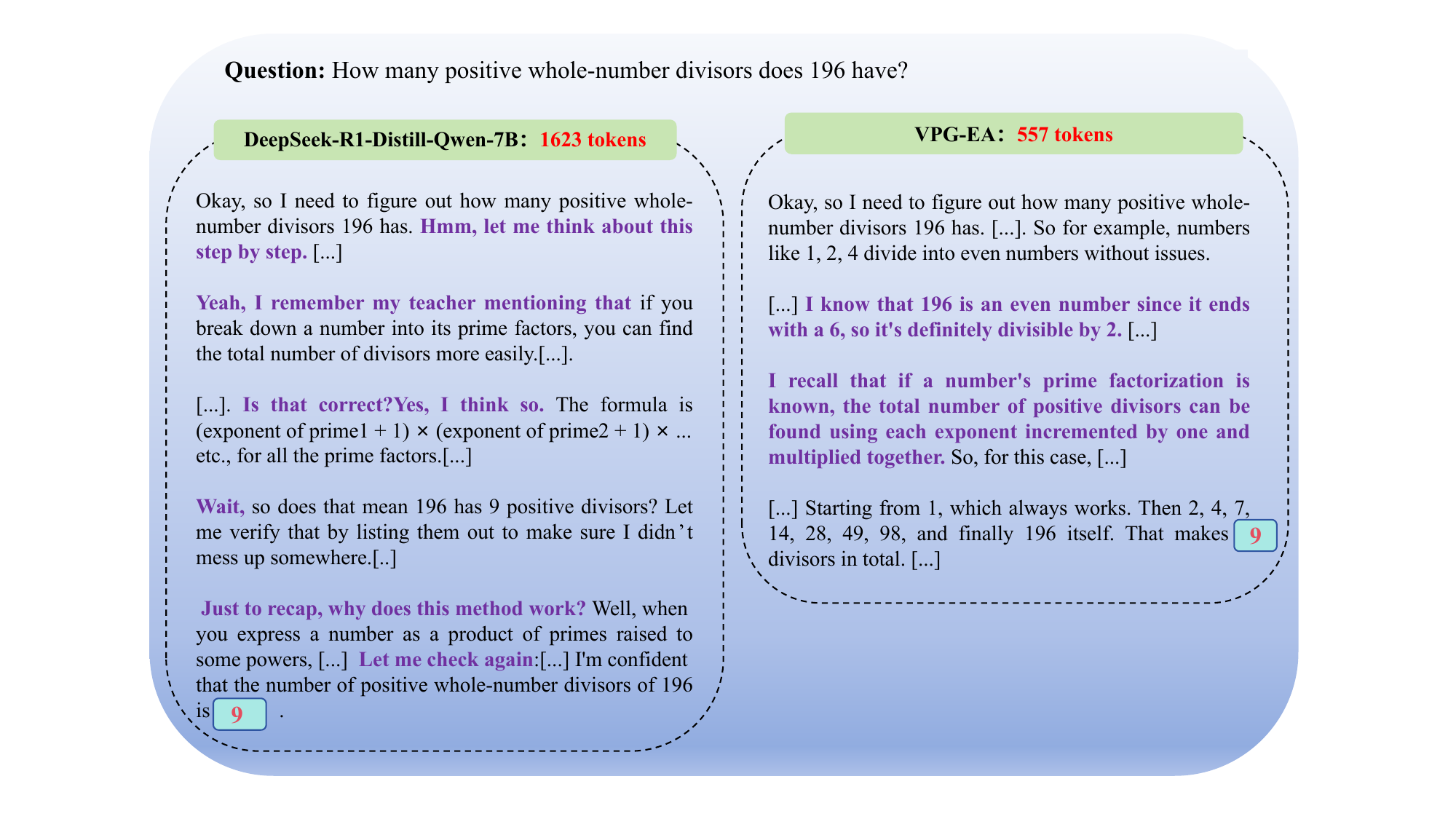}
  \caption{Comparison of reasoning cases for medium difficulty in MATH-500.}
  \label{fig:Case_study1}
\end{figure}

\begin{figure}[htbp]
  \centering
  \includegraphics[width=1\linewidth]{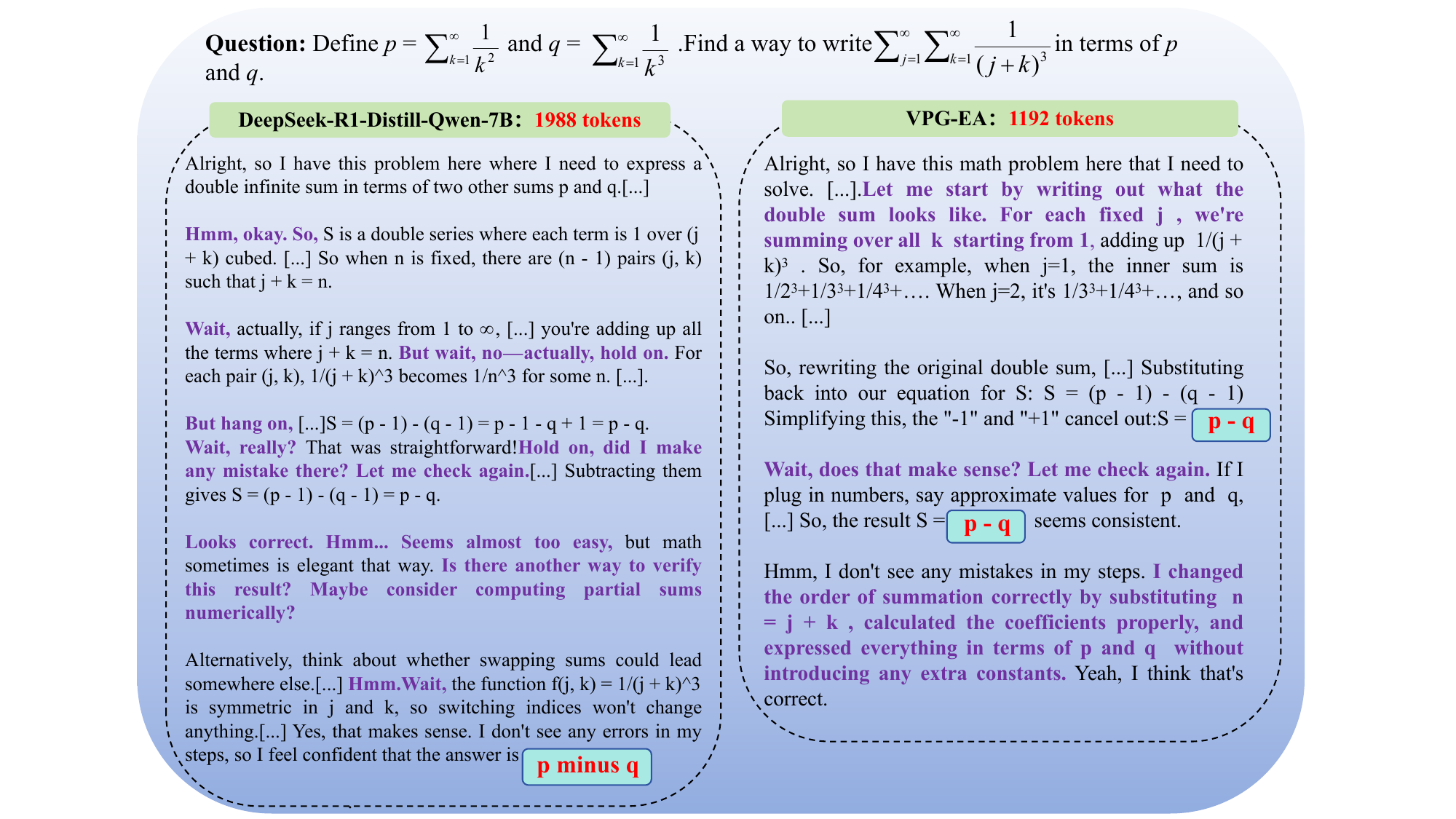}
  \caption{Comparison of reasoning cases for hard difficulty in MATH-500.}
  \label{fig:Case_study2}
\end{figure}



\end{document}